\documentclass[10pt,twocolumn,letterpaper]{article}
 \pdfoutput=1 
\usepackage{iccv}
\usepackage{times}
\usepackage{epsfig}
\usepackage{graphicx}
\usepackage{amsmath}
\usepackage{amssymb}
\usepackage[accsupp]{axessibility}
\usepackage{pdfpages}

\usepackage{subfiles}
\usepackage{subcaption}
\usepackage{algorithm}
\usepackage{algpseudocode}

\makeatletter
\newcommand\notsotiny{\@setfontsize\notsotiny\@vipt\@viipt}

\makeatother

\usepackage{xcolor}
\PassOptionsToPackage{hyphens}{url}\usepackage[pagebackref=true,breaklinks=true,letterpaper=true,colorlinks,
  citecolor=citecolor,bookmarks=false]{hyperref}
\definecolor{citecolor}{RGB}{34,139,34}

\usepackage[sort,nocompress]{cite}

\iccvfinalcopy

\ificcvfinal\fi

\begin{document} 
\title{On the Importance of Distractors for Few-Shot Classification}
 
\author{
{Rajshekhar Das$^{1}$ \qquad Yu-Xiong Wang$^{2}$ \qquad  Jos\'{e}  M.F. Moura$^1$} \\
{$^1$Carnegie Mellon University \qquad $^2$University of Illinois at Urbana-Champaign}\\
{\tt\small rajshekd@andrew.cmu.edu \qquad yxw@illinois.edu \qquad moura@andrew.cmu.edu}
}

\maketitle
\ificcvfinal\thispagestyle{empty}\fi
 
\begin{abstract}
Few-shot classification aims at classifying categories of a novel task by learning from just a few (typically, 1 to 5) labelled examples. An effective approach to few-shot classification involves a prior model trained on a large-sample base domain, which is then finetuned over the novel few-shot task to yield generalizable representations. However, task-specific finetuning is prone to overfitting due to the lack of enough training examples. To alleviate this issue, we propose a new finetuning approach based on contrastive learning that reuses unlabelled examples from the base domain in the form of distractors. Unlike the nature of unlabelled data used in prior works, distractors belong to classes that do not overlap with the novel categories. We demonstrate for the first time that inclusion of such distractors can significantly boost few-shot generalization. Our technical novelty includes a stochastic pairing of examples sharing the same category in the few-shot task and a weighting term that controls the relative influence of task-specific negatives and distractors. An important aspect of our finetuning objective is that it is agnostic to distractor labels and hence applicable to various base domain settings. Compared to state-of-the-art approaches, our method shows accuracy gains of up to $12\%$ in cross-domain and up to $5\%$ in unsupervised prior-learning settings. Our code is available at \url{https://github.com/quantacode/Contrastive-Finetuning.git}
\end{abstract}

\section{Introduction}
 
\begin{figure}[ht]
\centering 
         \includegraphics[width=.9\linewidth]{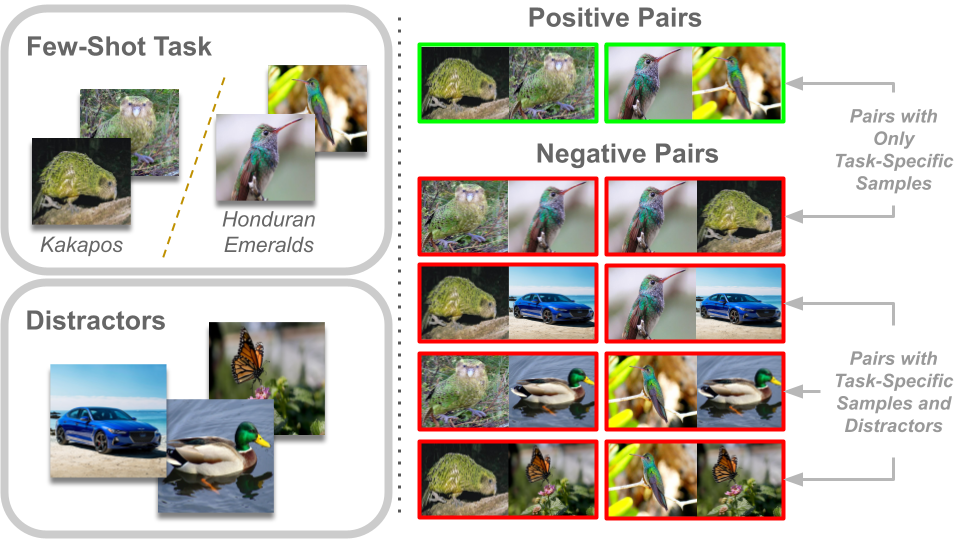}
\caption{\textbf{Classification of Kakapos vs. Honduran Emeralds with just few examples per class and many distractors}: The idea is to leverage unlabelled data in the form of \textit{distractors} that need not be semantically related to the classes in the few-shot task. The hope is that by pairing distractors and task samples as negatives (bottom six red boxes) and encouraging greater dissimilarity between such pairs, image representations of the two classes, Kakapos and Honduran Emeralds, will be pushed farther away. This would ultimately lead to better classification.}
\label{fig:teaser}
\end{figure}

The ability to learn from very few examples is innate to human intelligence. In contrast, large amounts of labelled examples are required by modern machine learning algorithms to learn a new task. This limits their applicability to domains where data is either expensive to annotate and collect or simply inaccessible due to privacy concerns. To overcome this limitation,   few-shot classification has been proposed as a generic framework for learning to classify with very limited supervision \cite{finn2017model,Snell2017PrototypicalNF, Lake2015HumanlevelCL, Koch2015SiameseNN}.
Under this paradigm, most approaches leverage prior knowledge from a (labelled) \textit{base} domain 
to solve a novel task by either finetuning-based transfer~\cite{tian2020rethink, Dhillon2020A} or meta-learning~\cite{Snell2017PrototypicalNF, finn2017model,NIPS2016Match6385,sung2018learning, Ravi2017OptimizationAA, Gidaris2018DynamicFV,garcia2018fewshot}. In particular, when the base and novel domains are related, the hope is that representations learnt in the base domain can be generalized to novel tasks, thus facilitating {\em positive} knowledge transfer. 

While the above paradigm is effective for tasks that can leverage large datasets like ImageNet \cite{ILSVRC15} as the related base domain, for others, such as rare species classification \cite{van2018inaturalist} or medical image classification\cite{WangChest2017}, acquiring necessary prior knowledge  can be exceedingly difficult due to the absence of a related base domain with labelled data. To relax such data requirements, recent techniques explore alternative ways such as unsupervised learning \cite{khodadadeh2021unsupervised,hsu2018unsupervised} or cross-domain learning~\cite{crossdomainfewshot,afrasiyabi2020associative,Gidaris2018DynamicFV, phoo2021selftraining} to obtain representations useful for novel tasks. In the absence of labelled base data, approaches like \cite{hsu2018unsupervised,khodadadeh2019unsupervised,khodadadeh2021unsupervised} seek to benefit from  self-supervised representation learning over unlabelled data in a related domain. In a more challenging scenario where related base data is hard to obtain, cross-domain techniques \cite{crossdomainfewshot, Doersch2020CrossTransformersSF,tian2020rethink} exploit representations learnt in other domains that do not have the same task characteristics as the novel tasks.

Although the issue of learning a good prior representation remains a core focus in few-shot classification, it addresses only a part of the problem. In this work, we investigate the other important aspect, \ie, {\em effective finetuning specific to the novel task}. Our main motivation comes from recent findings \cite{guo2020broader,afrasiyabi2020associative, Dhillon2020A} that demonstrate the outperformance of simple finetuning over more sophisticated prior learning techniques such as meta-learning. Despite its effectiveness, we suspect that finetuning might still suffer from overfitting as a consequence of small training set in a few-shot task. To alleviate this situation, we propose to {\em leverage additional unlabelled data exclusive to the task}. Such datapoints are referred to as \textit{distractors}.  For instance, in the case of classifying  Honduran Emeralds and Kakapos (rare species of birds), examples of butterflies, cars or ducks can serve as distractors (Fig.~\ref{fig:teaser}). By the virtue of its task-exclusivity, distractors can be obtained from various data-abundant domains with categories that could be semantically unrelated to novel task categories. However, in this work, we restrict ourselves to just the base data as a source for distractors. This allows us to efficiently reuse the data under standard settings and directly compare with prior works.

To this end, we pose the imminent question -- \textit{Can distractors improve few-shot generalization?} The answer is, somewhat surprisingly, yes. To elucidate how, we propose  \textit{ConFT}, a simple \textbf{f}ine\textbf{t}uning method based on a \textbf{con}trastive loss that contrasts pairs of the same class against those from different classes. We show that with a few simple but crucial modifications to the standard contrastive loss, distractors can be incorporated to boost generalization.  We hypothesize that in the absence of extensive in-domain supervision for prior experience, distractor-aware finetuning can yield non-trivial gains. Towards the design of the loss function, we adopt an {\em asymmetric} construction of similarity pairs to ensure that {\em distractors contribute only through different-class pairs}. Our key insight here  is two-fold -- 1) generalization in contrastive learning can be influenced by not only same-class but also different-class pairs; 2) construction of different-class pairs is extremely flexible in that it can include samples from task-specific as well as task-exclusive categories. As a test of generality, we study the effect of our finetuning approach in conjunction with two different prior learning setups, namely, cross-domain and unsupervised prior learning. {\bf Our contributions} are as follows.
\begin{itemize}
\item We propose contrastive finetuning, \textit{ConFT}, a novel finetuning method for transfer based few-shot classification.
\item We show how distractors can be incorporated in a contrastive objective to improve few-shot generalization.
\item The proposed method outperforms state-of-the-art approaches by up to $12$ points in the cross-domain few-shot learning and up to  $5 $ points  in  unsupervised prior learning settings.
\end{itemize}

\section{Related Work}
\subsection{Few-Shot Classification}
Modern algorithms for few-shot classification are predominantly based on meta-learning where the goal is to quickly adapt to novel tasks. These approaches can be broadly classified into three categories: initialization based \cite{finn2017model, nichol2018reptile, rusu2019meta, ravi2017optimization, munkhdalai2017meta}, hallucination based~\cite{hariharan2016low, antoniou2018data, wang2018low}, and metric-learning based~\cite{bromley1993signature,NIPS2016Match6385, Snell2017PrototypicalNF, sung2018learning, garcia2018fewshot, Koch2015SiameseNN} methods. Despite the growing interest in sophisticated meta-learning techniques, recent works \cite{tian2020rethink,Dhillon2020A,afrasiyabi2020associative,chen2019closer} have demonstrated that even simple finetuning based transfer learning \cite{5288526,NIPS2014375c7134, Kornblith2019DoBI, Ge2017CVPR, guo2018spottune} can outperform them. Such baselines usually involve cross-entropy training over the base categories followed by finetuning over a disjoint set of novel classes. Following these results, we further the investigation of finetuning for few-shot classification.

\textbf{Cross-Domain Few-Shot Classification:} A number of recent works \cite{crossdomainfewshot,guo2020broader,dvornik2019diversity,afrasiyabi2020associative,oh2021boil,ouali2020spatial,Tseng2020Cross-Domain, Ryu2020MetaPerturbTR} have been proposed to address the cross-domain setup where base and novel classes are not only disjoint but also belong to different domains. Interestingly, \cite{chen2019closer} demonstrated that in this setup too, finetuning based transfer approaches outperformed popular meta-learning methods by significant margins. Following that, \cite{crossdomainfewshot} proposed to learn feature-wise transformations via meta-learning to improve few-shot generalization of metric-based approaches. While in standard finetuning, the embedding model is usually frozen to avoid overfitting, recent works like \cite{afrasiyabi2020associative, guo2020broader} have shown that frozen embeddings can hinder few-shot generalization. In this work, we build upon these developments to propose a more effective finetuning method over the entire embedding model.

In the context of learning from heterogeneous domain, \cite{triantafillou2019metadataset} introduced a benchmark for multi-domain few-shot classification. This benchmark has been adopted by some recent works \cite{Saikia2020OptimizedGF, Doersch2020CrossTransformersSF, Chen2020ANM, Majumder2021RevisitingCL}. While multiple base domains can alleviate cross-domain learning, we test our approach on a more challenging setup \cite{crossdomainfewshot} that only involves a single base domain. Recent works used \cite{crossdomainfewshot} as a benchmark to evaluate the importance of representation change \cite{oh2021boil} and spatial contrastive learning \cite{ouali2020spatial} in cross-domain few-shot classification.  Another related work \cite{Tseng2020Cross-Domain} leveraged unlabelled data from the novel domain in addition to few-shot labelled data to improve the task performance in a similar benchmark \cite{guo2020broader}. In contrast to \cite{Tseng2020Cross-Domain}, we operate under a limited access to novel domain data, \ie only the few-shot labelled data.

\textbf{Unlabelled Data in Few-Shot Classification:} Our use of unlabelled data in the form of distractors is inspired from cognitive neuroscience studies  \cite{liang-etal-2018-distractor} describing the effect of visual distractors on learning and memory.  Prior works that use additional unlabelled data for few-shot classification include \cite{Su2020When, gidaris2019boosting,Bukchin2020FinegrainedAC, Wang-2016-26050, ren2018metalearning,NEURIPS2019bf25356f}. Complementary to \cite{Su2020When,gidaris2019boosting,Bukchin2020FinegrainedAC} that exploit unlabelled data via self-supervised objectives in the prior learning phase, we use unlabelled data specifically for task-specific finetuning. Nonetheless, combining both perspectives could yield further benefits and is left for future work.

More related approaches \cite{ren2018metalearning,NEURIPS2019bf25356f} combined heterogeneous unlabelled data, \ie, task-specific data and distractors, in a semi-supervised framework. Our distractor-aware finetuning differs from these works in two important ways: our few-shot classification is strictly inductive in that we do not use unlabelled data specific to the task, and our method leverages distractors instead of treating them as interference that needs to be masked out. The most relevant methods \cite{afrasiyabi2020associative,Ge2017CVPR}, like us, reused the base (or source) domain as a source for additional data. The key difference, however, is that their success relies on effective alignment of the base and novel classes, whereas we benefit from contrasting the two. While the importance of distractor-aware learning has been investigated in the context of object detection~\cite{Zhu18ECCV,PosseggerCVPR}, their benefit to few-shot generalization has not been studied before. 

Recently,  \cite{hsu2018unsupervised,khodadadeh2019unsupervised,khodadadeh2021unsupervised} have studied few-shot classification in the context of unsupervised prior-learning where the base data is unlabelled. In this work, we evaluate the benefit of contrastive finetuning under this setting and and compare it to existing methods.

\subsection{Contrastive Learning}
Contrastive learning yields a similarity distribution over data by comparing pairs of different samples \cite{pmlr-v97-saunshi19a}. Recently, contrastive learning \cite{NIPS20166b180037, 1640964,pmlr-v9-gutmann10a, pmlr-v2-salakhutdinov07a} based methods have emerged as the state of the art for supervised \cite{NEURIPS2020SupCond89a66c7, wu2018improving, Kamnitsas2018SemiSupervisedLV} and self-supervised \cite{chen2020simple,He2020CVPR, tian2019contrastive, Hnaff2020DataEfficientIR, hjelm2018learning, wu2018unsupervised, Oord2018RepresentationLW} representation learning. While the supervised approaches primarily exploit ground-truth labels to construct same-class pairs, self-supervised techniques leverage domain knowledge in the form of data augmentation to generate such pairs. As a special case, \cite{NEURIPS2020SupCond89a66c7} maximized the benefit by integrating both forms of contrastive losses into a single objective. In this work, we use a modified version of the supervised contrastive loss when more than one labelled example is available per category. However, in the extreme case of 1-shot classification, it switches to self-supervised contrastive learning.
Recent works such as \cite{Doersch2020CrossTransformersSF,ouali2020spatial} also explored contrastive learning in the context of few-shot classification. While they use contrastive objectives at the \emph{prior-learning} stage to learn a general-purpose representation solely on the base domain, our method uses a contrastive objective at \emph{finetuning} to improve the \emph{downstream-task-specific} representation directly on the target domain task with base domain data as distractors. As a design choice, we adopt the contrastive loss over other losses like cross-entropy since it allows us to leverage distractor data that does not belong to the novel categories but improves generalization. 

\section{Our Approach}
 \label{sec:conft}
 
 \begin{figure*}[t]
\centering 
         \includegraphics[width=.7\linewidth]{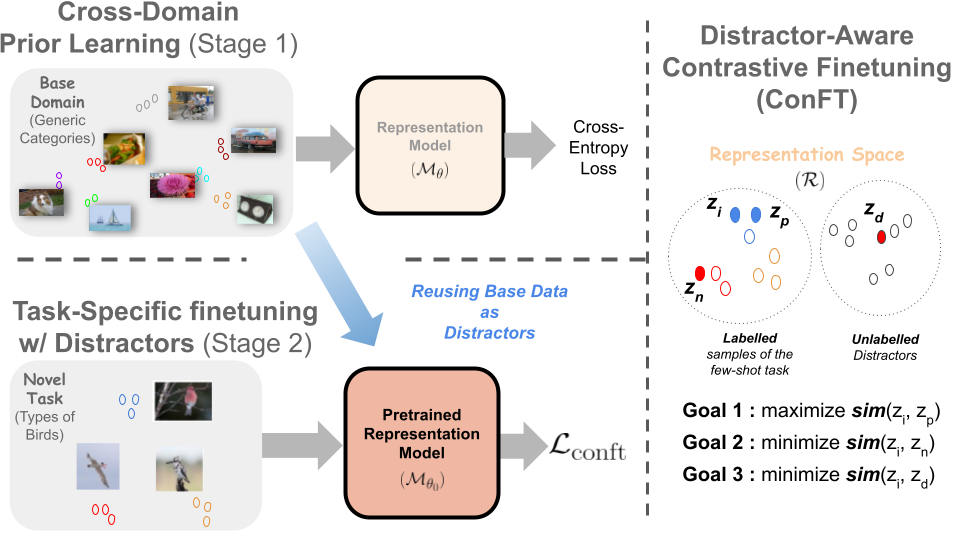}
\caption{\textbf{Contrastive Finetuning in Cross-Domain Few-Shot Learning:} Our contrastive finetuning approach to few-shot classification comprises of two stages: 1) The \textit{prior learning} stage trains a representation model on labelled (under cross-domain settings) base data using a cross-entropy loss; 2) The pretrained representation model is then finetuned over \textit{task-specific} samples as well as \textit{distractors} using a contrastive loss. For each task sample $z_i$, the contrastive objective (\textit{right}) maximizes a similarity score, \textit{sim}, over same-class pairs while minimizing it over other pair types. In the absence of enough labelled examples, distractors can improve classification by pushing apart task-specific clusters (here, different classes of birds).}
\label{fig:pipeline}
\end{figure*}

 To achieve the goal of few-shot generalization, our contrastive finetuning method, ConFT, optimizes for two simultaneous objectives. First, it aims to bring  task-specific samples that share the same class close to each other; and second, it strives to push apart samples that belong to different classes. This two-fold objective can lead to compact clusters that are well separated amongst each other. In the following sections, we first introduce some notations that we then use to formally describe our approach.
An overview of our method is presented in Fig.~\ref{fig:pipeline}.

 \subsection{Preliminaries}
Consider an input space $\mathcal{X}$ and a categorical label set $\mathcal{Y}=\{c_1, \ldots, c_M\}$ where each of the $M$ classes is represented via one-hot encoding. A representation space $\mathcal{R} \subset \mathbb{R}^r$ of the input is defined by the composition of an augmentation function $\mathcal{A}:\mathcal{X} \to \mathcal{X}$ and a representation model $\mathcal{M}_\theta : \mathcal{X}\to\mathcal{R}$, parameterized by $\theta$.  The augmentation function is a composition of standard image transformations such as random cropping, color jittering, horizontal flipping \etc.
Given a small number $K$, a few-shot classification task $\tau$ can be defined as the collection of a support set, $\tau_\text{supp} = \{(x_i, y_i)|~x_i\in\mathcal{X},~y_i\in\mathcal{Y}, i\in I_\text{supp}\}$  with $K$ examples per class, and a query set, $\tau_q = \{\tilde{x}_j|~\tilde{x}_j\in\mathcal{X}, j\in I_q\}$ sampled from the same (but unobserved) classes. Here, $I_\text{supp}$ and $I_q$ are the collection of indices for the support and query sets, respectively.  
The few-shot classification goal is to leverage the support set to obtain a classifier for the query samples. In this case, the classifier is constructed over the representation model obtained via {\em contrastive finetuning} of a prior model, $\mathcal{M}_{\theta_0}$ over $\tau_\text{supp}$.
\subsection{The ConFT Objective}
\label{subsec:conft}
A key component of the ConFT objective is that it includes unlabelled samples, \textit{distractors}, to improve few-shot generalization. Formally, a distractor set, $S_\text{dt} = \{x_i|x_i\in\mathcal{X}, i\in I_\text{dt}\}$, drawn from a domain $D : \mathcal{X}\times\mathcal{Y}_D$ together with the task-specific support set $\tau_\text{supp}$, constitutes the training data for few-shot learning. Here, the distractor class set $\mathcal{Y}_D$ is assumed to be {\em task-exclusive}, $\mathcal{Y}_D \cap \mathcal{Y}=\emptyset$. Starting with a support set example $i$ (a.k.a anchor), we first construct an anchor-negative index set, $N(i) =  \{p\in I_\text{supp}|~ y_i\neq y_p\}$, and an anchor-positive index set $P(i)$
such that $y_p=y_i, \forall p\in P(i)$. Samples indexed by $N(i)$ are treated as negatives within the task, whereas those indexed by $I_\text{dt}$ act as negatives exclusive to the task. Finally, we define our contrastive loss that uses a $l_2$-normalized representation $z\in \mathbb{R}^r$ as follows  
\begin{align} 
&\mathcal{L}_\text{conft}(\theta) = - \frac{1}{|I_\text{supp}|} \sum\limits_{i \in I_\text{supp}} \frac{1}{|P(i)|}\sum\limits_{p\in P(i)} \log l_{ip}, \nonumber\\
&l_{ip} = \frac{\exp(\frac{z_i \cdot z_p}{\gamma})}{\exp(\frac{z_i \cdot z_p}{\gamma})+ \sum\limits_{n\in N(i)} \exp(\frac{z_i \cdot z_n}{\gamma})  + \sum\limits_{d\in I_\text{dt}}\exp(\frac{z_i \cdot z_d}{\gamma})}
\label{eq:conft_vanila}
\end{align}
where $\gamma$  is a temperature  hyper-parameter.
The finetuning objective is simply the minimization of $\mathcal{L}_\text{conft}$ to yield optimal parameters $\theta_\tau$ specific to task $\tau$. To classify the query samples, we construct a nearest-mean classifier  \cite{pmlr-v97-saunshi19a, 46678, dvornik2019diversity} atop the updated representation $\mathcal{M}_{\theta_\tau}$. The class-specific weight vectors are computed as an average over the representations of $K$ support examples pertaining to that class. The $j^\text{th}$ query sample is then assigned to the class whose weight vector has the largest cosine similarity (and hence, nearest in the Euclidean sense) with the query representation. We use the accuracy of this classifier to compare various baselines in the experiment section. 

\textbf{Construction of Anchor-Positive Set $P(i)$:} To construct an anchor-positive set, we randomly pair task-samples belonging to the same class with no sample occurring in more than one pair. In each pair, if one is assigned to be the anchor, the other acts as its positive. As an example, in a $5$-way $4$-shot task, our stochastic pair construction will result in 10 pairs where each of the $5$ classes has $2$ pairs. In the case when the number of shots is odd, we omit one sample from each class to allow even pairing. The omission is, however, not an issue in the overall scheme of finetuning where multiple steps of gradient descent optimization ensures that eventually each sample gets to participate with equal chance.  In the special case $1$-shot learning, anchor-positive sets are constructed similar to \cite{chen2020simple} using augmentation $\mathcal{A}$.

\subsection{Relative Importance of Anchor-Negatives}
Given the loss formulation of ~\eqref{eq:conft_vanila}, 
both task-specific (few-shot) and task-exclusive (distractor) anchor-negatives influence the loss proportionate to their respective batch sizes. While the batch size of task-specific negatives $N(i)$ is upper bounded by the number of ways M and the number of shots K, the batch size of distractors can be made as large as that of the domain itself, \ie, $|D|$. In standard contrastive learning paradigms with only task-specific and no task-exclusive training examples, large batch sizes of negatives are known to be beneficial for downstream task performance. However, in our case where both types of negatives exist, naively increasing distractor batch size can be counterproductive (shown in the supplementary). We suspect that too many distractors might overshadow the effect of task-specific negatives that can be more crucial for generalization. Also, the effect might vary according to the proximity of distractors with respect to task samples in the representation space. Nonetheless, there is a need to balance the undue influence of distractors by adjusting the batch sizes. To avoid an extensive search for an optimal batch size specific to the distractor domain, we propose a domain-agnostic weighting scheme for the anchor-negatives proportional to their batch sizes as follows
\begin{align}
&l_{ip} =\notag\\
& \frac{\exp(\frac{z_i \cdot z_p}{\gamma})}{\exp(\frac{z_i \cdot z_p}{\gamma})+ \alpha\sum\limits_{n\in N(i)} \exp(\frac{z_i \cdot z_n}{\gamma})  + (2-\alpha)\sum\limits_{d\in I_\text{dt}}\exp(\frac{z_i \cdot z_d}{\gamma})}, \label{eq:conft}\\
&\alpha = 2\frac{|I_\text{dt}|}{|N(i)| +  |I_\text{dt}|}\nonumber 
\end{align}
We found that this simple weighting scheme  makes the few-shot performance robust to batch size variations and also improves the overall performance (see the supplementary).
 
\section{Prior Learning and Distractors}
Thus far we have assumed the access to a distractor set, $S_\text{dt}$  and a prior model $\mathcal{M}_\theta$. In this section, we describe how to obtain them and how distractors boost generalization. Recall that our goal is to achieve few-shot generalization by finetuning a prior model over the few-shot task. However, due to the scarcity of task-specific labelled examples, a reasonably strong prior encoded in the model parameters $\theta_0$ is crucial for preventing overfitting, especially when using high-capacity models like neural networks. We next describe two different ways of learning such a prior that can serve as a good initialization for subsequent finetuning. 

\subsection{Types of Prior Learning}
\textbf{Cross-Domain Learning:} In the cross-domain setup, we are provided a labelled dataset, $D_l = \{(x_i, y_i)|x_i\in \mathcal{X}_\text{sc}, y_i \in \mathcal{Y}_\text{sc}\}_{i=1}^{|D_l|}$ drawn from a source domain $\mathcal{X}_\text{sc}\times\mathcal{Y}_\text{sc}$, such that the categorical label set $\mathcal{Y}_\text{sc}$ is disjoint from novel categories $\mathcal{Y}$. The key characteristic of this setup is that the distribution of $M$-way $K$-shot tasks, if constructed out of $D_l$, will be significantly different from novel tasks in the target domain. Such distribution shift could arise due to difference in task granularity (\eg, coarse-grained vs. fine-grained) or shift in input distribution or both. In this work, we consider the case where the shift in task granularity is notably more than the input distribution. Towards the goal of learning a reasonably strong prior, we adopt a simple objective that minimizes cross-entropy loss over all categories in $D_l$. During finetuning, the distractors are sampled from $D_l$, thus, naturally satisfying the non-overlapping categories assumption with respect to novel tasks. 

\textbf{Unsupervised Prior Learning: }
For unsupervised prior learning, we are given an unlabelled dataset, $D_u = \{v_i\}_{i=1}^{|D_u|}$ drawn from a source domain $\mathcal{X}_\text{su}\times\mathcal{Y}_\text{su}$, such that the corresponding labels in $\mathcal{Y}_\text{su}$ are unobserved.
While there are no explicit assumptions about the task distribution gap in this setting, the strength of the learnt prior is likely to be more reasonable when the distribution gap is small. To learn a suitable prior using $D_u$, we use the SimCLR loss \cite{chen2020simple} as a form of self-supervised objective. Our choice of this objective over others  \cite{He2020CVPR} was based on its superior performance found in our preliminary experiments. Priors learnt via self-supervised contrastive objectives on large base datasets (like, ImageNet) have been shown to transfer well to many-shot downstream tasks. In this work we show that such objectives are effective even with smaller base datasets and few-shot downstream tasks. In our experiments, priors learnt in this way already outperform state-of-the-art approaches \cite{khodadadeh2019unsupervised,khodadadeh2021unsupervised,hsu2018unsupervised} that are then further improved by our proposed contrastive finetuning. In this setup, we use $D_u$ as the source for distractors where the assumption of non-overlapping categories is satisfied with high probability, provided the base dataset is relatively large and encapsulates a wide variety of categorical concepts.

\subsection{Distractor-Aware Generalization}
\label{sec:gen_explain}
The most important and perhaps surprising aspect of our method is that distractors, despite being drawn from unrelated (to novel task) categories, can improve generalization. To understand the underlying mechanism, we propose to measure the change in quality of task-specific representation before and after finetuning. Particularly, given a few-shot task with $M$ classes, we define the subset of query samples, $I^c_q \subset I_q$  that share the same class\footnote{Note that the query class labels are considered only for analysis purposes. In practice, they are not observed.} and two other quantities -- cluster spread $u^q_\text{spread}$ and cluster-separation $u^q_\text{sep}$ that measure the degree of clustering in the representation space. Specifically,

\begin{align}
u^q_\text{spread}(\theta_t) = \frac{1}{M}\sum\limits_{m=1}^M \sum\limits_{\begin{subarray}{c}i\in I_q^{c_m} \\ j \in I_q^{c_m}\setminus\{i\}\end{subarray}}    (1-z_i \cdot z_j), \\
u^q_\text{sep}(\theta_t) =\frac{1}{M}\sum\limits_{m=1}^M \sum\limits_{\begin{subarray}{c}i\in I_q^{c_m},  \\j\in I_q\setminus I_q^{c_m} \end{subarray}}   (1- z_i \cdot z_j),
\label{eq:gen_mech}
\end{align} 
where $\theta_t$ are the parameters of the representation model after $t$ finetuning epochs. 
For each of the above quantities, we define the change, $\delta^q_* (t)= u^q_* (\theta_t) - u^q_* (\theta_0)$,  and relative change, $\delta^\text{rel,q}_*(t) = \frac{\delta^q_*(t)}{\kappa(\theta_0)}$ where, the subscript can be \textit{sep} or \textit{spread} and division by a fixed value, $\kappa(\theta_0)$ ensures scale invariance. Finally, to quantify  generalization within a given target domain, we define the average relative change, $\mathbb{E}_\tau~[\delta^{\text{rel,q}}_*(t)]$ over a large number of tasks sampled from that domain. The average relative change can also be defined for support examples by simply swapping superscript `q' with `s'. Also, in practice, we use $u^s_\text{sep}(\theta_0)$ as the fixed value for $\kappa(\theta_0)$ irrespective of the superscript or subscript. 

\subsection{A Multitask Variant of ConFT}
While our original objective \eqref{eq:conft} is agnostic to distractor supervision, finetuning in the cross-domain setting can further benefit from distractor labels. To that end, we introduce an auxiliary loss $\mathcal{L}_\text{mtce}$ 
during finetuning that minimizes the cross-entropy  between predicted probabilities and one-hot encodings of the ground-truth label averaged over the base data, $D_l$. This leads to a new multitask formulation 
 \begin{equation}
 \mathcal{L}_\text{mt-conft} = \mathcal{L}_\text{conft} + \lambda \mathcal{L}_\text{mtce},
 \label{eq:mtconft}
 \end{equation}
where we fix the relative weighting factor $\lambda=1$ in our experiments and use a cosine classifier \cite{chen2019closer} for $\mathcal{L}_\text{mtce}$. We found that this simple extension led to significant performance gains in some domains while marginal in others, depending on domain characteristics.

\section{Experiments}
 
Following sections first introduce some baselines (\S \ref{exp:baseline}) and present our main results  for contrastive finetuning in the cross-domain setup (\S \ref{exp:cdfsl}). Then, \S \ref{exp:fsg} elucidates the generalization mechanism of ConFT followed by  ablations in \S \ref{exp:ablation}. Finally, \S \ref{exp:ufsl} demonstrates the performance of our approach in the unsupervised prior learning setup.  

\textbf{Datasets and Benchmarks:}
We evaluate our proposed finetuning method in a variety of novel domains spanning across two different paradigms for prior learning. For cross-domain evaluations, we adopt the benchmark introduced by \cite{crossdomainfewshot} that comprises of Cars\cite{6755945}, CUB\cite{WelinderEtal2010}, Places \cite{7968387}, and Plantae \cite{van2018inaturalist} as the novel domains and \textit{mini}ImageNet \cite{Ravi2017OptimizationAA} as the base domain. Each dataset is split into \textit{train}, \textit{val} and, \textit{test} categories (please refer to the supplementary for details), where tasks sampled from the \textit{test} split are used to evaluate the few-shot performance in respective domains. We use the val splits for cross-validating the hyperparameters and the train split of \textit{mini}ImageNet as our base data. For experiments in unsupervised prior learning, we use the same train split of \textit{mini}ImageNet to learn a self-supervised representation that is then evaluated for few-shot performance on \textit{mini}ImageNet-test. We present additional results on Meta-Dataset \cite{triantafillou2019metadataset} in the supplementary.

\textbf{Backbone (Representation Model):} Following best practices in cross-domain few-shot learning, we adopt a ResNet10 \cite{He2016DeepRL} model for most of our experiments. In the unsupervised learning case, we use a four-layer CNN consistent with existing works except for a reduced filter size from $64$ to $20$ in the final layer. This modification was found to improve contrastive finetuning performance. 

\textbf{Optimization and Hyperparameters:}
In this work, we evaluate few-shot performance over $5$-way $1$-shot and $5$-way $5$-shot tasks with $15$ query samples, irrespective of the prior learning setup. For the contrastive finetuning, we use an ADAM \cite{Kingma2015AdamAM} optimizer with a suitable learning rate and early-stopping criteria. Our proposed method has a few hyperparameters such as the temperature ($\gamma$), learning rate, early-stopping criteria, and data augmentation ($\mathcal{A}$). However, recent studies \cite{oliver2018realistic} have highlighted that excessive hyperparameter tuning on large validation sets can lead to overoptimistic results in limited-labelled data settings like semi-supervised learning. Thus, we keep an extremely small budget for hyperparameter tuning. Among the mentioned hyperparameter, the one with the most number of parameters is the augmentation function $\mathcal{A}$. In this work, we do not tune $\mathcal{A}$ to any specific target domain. Instead, we use a fixed augmentation scheme introduced by \cite{chen2019closer} for the cross-domain setting  and  AutoAugment \cite{Cubuk2018AutoAugmentLA} for the unsupervised prior learning case. Please refer to the supplementary for a detailed summary of hyperparameters used in our experiments.
 
\subsection{Baseline Comparisons}
\label{exp:baseline}

\begin{table*}[htbp]
\notsotiny 
\centering
\begin{tabular}{cccccccccc}
\hline
\multicolumn{2}{c}{Finetuning Method}                                                       & \multicolumn{2}{c}{CUB}               & \multicolumn{2}{c}{Cars}              & \multicolumn{2}{c}{Places}            & \multicolumn{2}{c}{Plantae}           \\
Loss                                                                      & FT Type         & 1-shot            & 5-shot            & 1-shot            & 5-shot            & 1-shot            & 5-shot            & 1-shot            & 5-shot            \\ \hline
Cross-Entropy & fixed-BB (LC) \cite{chen2019closer}  & 39.77 $\pm$ 0.66 & 51.33 $\pm$ 0.70 & 33.99 $\pm$ 0.64 & 44.14 $\pm$ 0.70 & 44.53 $\pm$ 0.75 & 55.94 $\pm$ 0.69 & 37.07 $\pm$ 0.70 & 46.58 $\pm$ 0.69 \\
                                                                        & fixed-BB (CC) \cite{chen2019closer}  &       43.26 $\pm$ 0.76            &     62.87 $\pm$ 0.74              &    25.33 $\pm$ 1.85               &    50.40 $\pm$ 0.74               &       47.70 $\pm$ 0.76            &      69.48 $\pm$ 0.69             &   40.49 $\pm$ 0.77                &       56.64 $\pm$ 0.72            \\
                                                                          & FT-all (LC)  \cite{guo2020broader}   & 40.81 $\pm$ 0.75     & 61.82 $\pm$ 0.72     & 34.50 $\pm$ 0.67     & 55.63 $\pm$ 0.75     & 45.91 $\pm$ 0.77     & 68.73 $\pm$ 0.73     & 37.51 $\pm$ 0.71     & 58.33 $\pm$ 0.68     \\
                                                                          & FT-all (CC) \cite{guo2020broader}     & 44.30 $\pm$ 0.73     & 67.05 $\pm$ 0.69     & 36.79 $\pm$ 0.76     & 57.65 $\pm$ 0.76     & 49.10 $\pm$ 0.78     & 70.32 $\pm$ 0.72     & 40.31 $\pm$ 0.76     & 61.30 $\pm$ 0.75     \\ \hline
Contrastive              
                                                                          &-           & 43.42 $\pm$ 0.75     & 62.80 $\pm$ 0.76     & 35.19 $\pm$ 0.66     & 51.41 $\pm$ 0.72     & 49.56 $\pm$ 0.80     & 70.71 $\pm$ 0.68     & 40.39 $\pm$ 0.79     & 55.54 $\pm$ 0.69     \\
                                                                          & ConFT (ours)    & 45.57 $\pm$ 0.76     & 70.53 $\pm$ 0.75     &\textbf{ 39.11 $\pm$ 0.77    } & 61.53 $\pm$ 0.75     & \textbf{49.97 $\pm$ 0.86}     & 72.09 $\pm$ 0.68     & \textbf{43.09 $\pm$ 0.78 }    & 62.54 $\pm$ 0.76     \\
                                                                          & MT-ConFT (ours) & \textbf{49.25 $\pm$ 0.83}     & \textbf{74.45 $\pm$ 0.71     }& 37.36 $\pm$ 0.69     & \textbf{62.54 $\pm$ 0.72 }    &\textbf{ 49.94 $\pm$ 0.81}     & \textbf{72.71 $\pm$ 0.69}     & 41.82 $\pm$ 0.75     & \textbf{63.01 $\pm$ 0.74}    \\ \hline
\end{tabular}
\caption{\textbf{Baseline Comparisons.} Results on $1$-shot and $5$-shot tasks on the LFT benchmark \cite{crossdomainfewshot}. These results are obtained by averaging over 600 novel tasks, each consisting of $5$ classes and $15$ queries per class. We also present $95\%$ confidence intervals. The train split of the \textit{mini}ImageNet dataset is used as base data. Here,  FT-all denotes the case where the entire embedding model is finetuned. Other abbreviations -- BB: Backbone model (ResNet-10), LC: Linear Classifier, CC: Cosine Classifier with a multiplication factor of 10.}
\label{tab:baseline}
\end{table*}

We begin our evaluations by comparing various baselines for finetuning in Table \ref{tab:baseline}. These include two simple baselines (introduced in \cite{chen2019closer}) and two strong baselines (introduced in \cite{guo2020broader}).
While the simple baselines freeze the backbones, the others allow finetuning over the entire embedding model. Another key difference is that the simple baselines are evaluated using standard linear evaluation \cite{chen2019closer,chen2020simple}, whereas the rest are evaluated using nearest-mean classifiers. We compare the performance of all these baselines to our vanilla and multi-task (MT) versions of ConFT.  Following previous works, the learning rates for the simple baselines are kept at $0.01$, whereas for others (including ours), we use smaller learning rates ($0.005$ or $0.0005$). We observe that among the baselines, the cosine classifier based baseline, FT-all (CC), outperforms the linear classifier based FT-all (LC). However, both versions of our finetuning approach significantly outperform all baselines across various dataset and shot settings. 

\subsection{ConFT for Cross-Domain Prior Learning }
\label{exp:cdfsl}

\begin{table*}
\scriptsize 
\centering 
\begin{tabular}{ccccccc}
\hline
\multicolumn{2}{c}{Method}                &           & \multicolumn{4}{c}{1-shot}                                        \\ \cline{4-7} 
Prior Learning & Task Specific Finetuning & Backbone  & CUB            & Cars           & Places         & Plantae        \\ \hline
AAL \cite{afrasiyabi2020associative}           & arcmax               & ResNet18  & 47.25 $\pm$ 0.76   & -              & -              & -              \\
MN   \cite{NIPS2016Match6385}          & -                        & ResNet10  & 35.89 $\pm$ 0.51 & 30.77 $\pm$ 0.47 & 49.86 $\pm$ 0.79 & 32.70 $\pm$ 0.60 \\
MN w/ featTx  \cite{crossdomainfewshot} & -                        & ResNet10  & 36.61 $\pm$ 0.53 & 29.82 $\pm$ 0.44 & 51.07 $\pm$ 0.68 & 34.48 $\pm$ 0.50 \\
RN  \cite{sung2018learning}           & -                        & ResNet10  & 42.44 $\pm$ 0.77 & 29.11 $\pm$ 0.60 & 48.64 $\pm$ 0.85 & 33.17 $\pm$ 0.64 \\
RN w/ featTx \cite{crossdomainfewshot}  & -                        & ResNet10  & 44.07 $\pm$ 0.77 & 28.63 $\pm$ 0.59 & 50.68 $\pm$ 0.87 & 33.14 $\pm$ 0.62 \\
GNN \cite{garcia2018fewshot}           & -                        & ResNet10+ & 45.69 $\pm$ 0.68 & 31.79 $\pm$ 0.51 & 53.10 $\pm$ 0.80 & 35.60 $\pm$ 0.56 \\
GNN w/ featTx  \cite{crossdomainfewshot} & -                        & ResNet10+ & 47.47 $\pm$ 0.75 & 31.61 $\pm$ 0.53 & \textbf{55.77 $\pm$ 0.79 }& 35.95 $\pm$ 0.58 \\
MAML \cite{finn2017model}          &      -                    & Conv4     & 40.51 $\pm$ 0.08   & 33.57 $\pm$ 0.14   & -              & -              \\
ANIL  \cite{Raghu2020Rapid}         &     -                     & Conv4     & 41.12 $\pm$ 0.15   & 34.77 $\pm$ 0.31   & -              & -              \\
BOIL \cite{oh2021boil}          &       -                   & Conv4     & 44.20 $\pm$ 0.15   & 36.12 $\pm$ 0.29   & -              & -              \\
\hline
CE Training    & -                        & ResNet10  & 43.42 $\pm$ 0.75  & 35.19 $\pm$ 0.66  & 49.56 $\pm$ 0.80  & 40.39 $\pm$ 0.79  \\
CE Training    & ConFT (ours)             & ResNet10  & 45.57 $\pm$ 0.76  &\textbf{ 39.11 $\pm$ 0.77 } & 49.97 $\pm$ 0.86  & \textbf{43.09 $\pm$ 0.78}  \\
CE Training    & MT-ConFT (ours) & ResNet10  & \textbf{49.25 $\pm$ 0.83 } & 37.36 $\pm$ 0.69  & 49.94 $\pm$ 0.81  & 41.82 $\pm$ 0.75  \\ \hline\hline
               &                          &           &                &                &                &                \\

\multicolumn{2}{c}{Method}                &           & \multicolumn{4}{c}{5-shot}                                        \\ \cline{4-7} 
Prior Learning & Task Specific Finetuning & Backbone  & CUB            & Cars           & Places         & Plantae        \\ \hline
Baseline     \cite{chen2019closer}  & -                        & ResNet18  & 65.57$\pm$0.70     &                &                &                \\
Baseline ++  \cite{chen2019closer}   & -                        & ResNet18  & 62.04$\pm$0.76     &                &                &                \\
DiversityNCoop  \cite{dvornik2019diversity}& -                        & ResNet18  & 66.17$\pm$0.55     & -              & -              & -              \\
AAL   \cite{afrasiyabi2020associative}            & arcmax        & ResNet18  & 72.37 $\pm$ 0.89   & -              & -              & -              \\
BOIL    \cite{oh2021boil}       & -                        & ResNet12  & -              & 49.71 $\pm$ 0.28   & -              & -              \\
MN \cite{NIPS2016Match6385}            & -                        & ResNet10  & 51.37 $\pm$ 0.77 & 38.99 $\pm$ 0.64 & 63.16 $\pm$ 0.77 & 46.53 $\pm$ 0.68 \\
MN w/ featTx  \cite{crossdomainfewshot} & -                        & ResNet10  & 55.23 $\pm$ 0.83 & 41.24 $\pm$ 0.65 & 64.55 $\pm$ 0.75 & 41.69 $\pm$ 0.63 \\
RN   \cite{sung2018learning}          & -                        & ResNet10  & 57.77 $\pm$ 0.69 & 37.33 $\pm$ 0.68 & 63.32 $\pm$ 0.76 & 44.00 $\pm$ 0.60 \\
RN w/ featTx \cite{crossdomainfewshot}   & -                        & ResNet10  & 59.46 $\pm$ 0.71 & 39.91 $\pm$ 0.69 & 66.28 $\pm$ 0.72 & 45.08 $\pm$ 0.59 \\
GNN  \cite{garcia2018fewshot}  \cite{garcia2018fewshot}        & -                        & ResNet10+ & 62.25 $\pm$ 0.65 & 44.28 $\pm$ 0.63 & 70.84 $\pm$ 0.65 & 52.53 $\pm$ 0.59 \\
GNN w/ featTx \cite{crossdomainfewshot}  & -                        & ResNet10+ & 66.98 $\pm$ 0.68 & 44.90 $\pm$ 0.64 &\textbf{ 73.94 $\pm$ 0.67} & 53.85 $\pm$ 0.62 \\
MAML      \cite{finn2017model}     &     -                     & Conv4     & 53.09 $\pm$ 0.16   & 44.56 $\pm$ 0.21   & -              & -              \\
ANIL  \cite{Raghu2020Rapid}          &      -                    & Conv4     & 55.82 $\pm$ 0.21   & 46.55 $\pm$ 0.29   & -              & -              \\
BOIL  \cite{oh2021boil}         &       -                   & Conv4     & 60.92 $\pm$ 0.11   & 50.64 $\pm$ 0.22   & -              & -              \\ \hline
CE Training    & -                        & ResNet10  & 62.80 $\pm$ 0.76  & 51.41 $\pm$ 0.72  & 70.71 $\pm$ 0.68  & 55.54 $\pm$ 0.69  \\
CE Training    & ConFT (ours)             & ResNet10  & 70.53 $\pm$ 0.75  & 61.53 $\pm$ 0.75  & 72.09 $\pm$ 0.68  & 62.54 $\pm$ 0.76  \\
CE Training    & MT-ConFT (ours) & ResNet10  & \textbf{74.45 $\pm$ 0.71}  & \textbf{62.54 $\pm$ 0.72 } & 72.71 $\pm$ 0.69  & \textbf{63.01 $\pm$ 0.74}) \\ \hline\hline
\end{tabular} 
\caption{\textbf{Cross-Domain Few-Shot Classification Results.} We present the results with $95\%$ confidence intervals and highlight the best performing methods. The results are an average over $600$ tasks. Here, `-' denotes numbers not reported by previous works.}
\label{tab:cdfsl}
\end{table*}
In this section, we present our main results on cross-domain few-shot learning (see Table \ref{tab:cdfsl}). We compare our approach with various prior works on the LFT benchmark \cite{crossdomainfewshot}. We observe that overall our proposed approaches, ConFT and MT-ConFT, significantly outperform the best previous results in Cars (by\textbf{ $3$ to $12$ points}), Plantae (by \textbf{$7$ to $9$ points}) and CUB (by \textbf{$1.7$ to $2$ points})  domains. We also observe higher gains in the $5$-shot setting than the $1$-shot case, since more labelled examples can improve few-shot generalization. Further, we find that using the auxiliary loss (MT-ConFT) is more beneficial in the $5$-shot case. In fact, it performs worse than ConFT in the $1$-shot cases for Cars, Places, and Plantae. Such a degradation could be due to a misalignment between the self-supervised objective (to which ConFT boils down in the 1-shot case) and the auxiliary cross-entropy loss. In the Places domain, ``GNN w/ featTx" yields the best performance, whereas our approach outperforms the rest for the $5$-shot case. We suspect that the use of a more sophisticated model in``GNN w/ featTx", namely, graph neural net \cite{garcia2018fewshot} built on top of a ResNet-10 model, leads to a better cross-domain generalization when the domain gap is smaller. 
 
\subsection{Effect of Distractors on Generalization}
\label{exp:fsg}

\begin{figure}[htbp]
\centering 
\begin{subfigure}[b]{0.9\columnwidth}
\centering
\includegraphics[width=0.9\columnwidth]{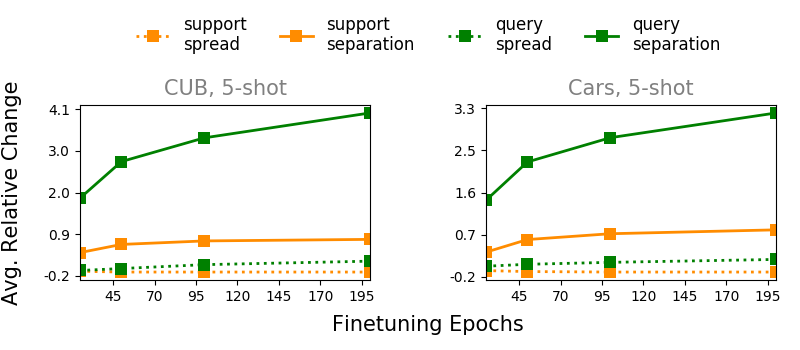} 
\end{subfigure}
\begin{subfigure}[b]{0.9\columnwidth}
\centering
\includegraphics[width=0.9\columnwidth]{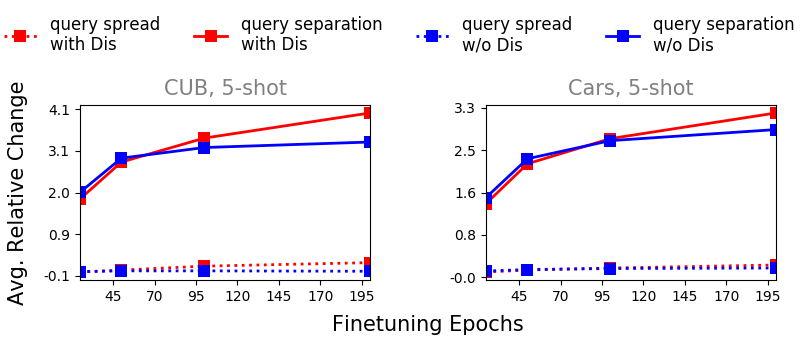}  
\end{subfigure} 
\caption{\textbf{Understanding Generalization in ConFT.} \textit{Top:} Average relative change in  cluster-spread and cluster-separation  of support and query samples as a function of finetuning epochs. \textit{Bottom:} Comparing the average relative change in cluster-spread and cluster-separation of only query samples under the presence and absence of distractors. The spread and separation quantities are averaged over 600 tasks for both \textit{top} and \textit{bottom}.}
\label{fig:gen}
\end{figure}

In this section, we investigate the central question -- \textit{How do distractors improve generalization?} We present two sets of plots in Figure \ref{fig:gen} that track the change in cluster-spread and cluster-separation as the finetuning progresses. In the first set, we plot the average relative change, $\mathbb{E}[\delta_\text{spread}^{\text{rel}}(t)]$  and $\mathbb{E}[\delta_\text{sep}^{\text{rel}}(t)]$ (see \S\ref{sec:gen_explain}) as a function of finetuning epochs, $t$ for both support and query samples in $2$ different settings, namely, CUB ($5$-shot) and Cars ($5$-shot). We observe that for support examples (yellow lines), cluster-spread decreases with increasing epochs while the cluster-separation increases. This is indeed what is expected for training datapoints (here, support examples) and serves as a sanity check. For query samples (green plot), on the other hand, both cluster-spread and separation increase with the progress in finetuning epochs. The key observation, however, is that cluster-separation increases to a much greater extent than the cluster-spread, thus improving overall discriminability between classes represented by these clusters. While the increase in cluster-separation hints towards the possible reason behind improved generalization, it is not clear how much of the improvement, if any, is a consequence of incorporating distractors. To delineate the effects of distractors from the contrastive loss itself, we present the second set of plots that compare the average relative change in \textit{query} cluster spread and separation under the presence  (red line)  and absence (blue line)  of distractors for the same data settings. We observe that with increasing finetuning epochs the gap between cluster-separation and cluster-spread widens to a larger extent in the presence of distractors than in their absence. This leads to our conclusion that \textbf{distractors help generalization by increasing task-specific cluster separation.}  

\subsection{Ablations}
\label{exp:ablation}
\begin{table}
\scriptsize  
\centering 
\begin{tabular}{ccccc|cc}
\hline
\multicolumn{2}{c}{Anchor-Positives}                                                                                                                                 & \multicolumn{3}{c|}{Anchor-Negatives}                                                   & \multicolumn{2}{c}{Accuracy}        \\
SPC &T-Pos &T-Neg &D-Neg&W &CUB, 5-shot &Cars, 5-shot\\
\hline
                                                                         & $\checkmark$                                                                              & $\checkmark$                                              & $\checkmark$ & $\checkmark$ & $69.61_{\pm 0.68}$                                        & $61.01_{\pm 0.74}  $                                       \\
$\checkmark$                                                             &                                                                                           &                                                           & $\checkmark$ &              & $70.16_{\pm 0.70} $                                      & $57.42 _{\pm 0.80} $                                         \\
$\checkmark$                                                             &                                                                                           & $\checkmark$                                              &              &              & $67.44 _{\pm 0.71}$                                        & $59.00 _{\pm 0.73}$                                         \\
$\checkmark$                                                             &                                                                                           & $\checkmark$                                              & $\checkmark$ &              & $70.26 _{\pm 0.68}$                                        & $60.58 _{\pm 0.77}$                                         \\
$\checkmark$                                                             &                                                                                           & $\checkmark$                                              & $\checkmark$ & $\checkmark$ & $70.53 _{\pm 0.75} $                                       & $61.53 _{\pm 0.75}$                                         \\ \hline
\end{tabular}
\caption{\textbf{Ablation 1.} Novel task performance with various types of anchor-positives and anchor-negatives. Here, SPC: Stochastic Pair Construction, T: Task, D: Distractor, W: relative weighting.}
\label{tab:abpn}
\end{table}

\begin{table}
\centering
\small 
\begin{tabular}{cccc|cc}
\hline
\multicolumn{4}{c|}{Distractor Domain Size}                                                                            & \multicolumn{2}{c}{Accuracy}                                                                                \\
512          & 1024         & 2048         & \begin{tabular}[c]{@{}c@{}}38400\end{tabular} & \begin{tabular}[c]{@{}c@{}}CUB 5-shot\end{tabular} & \begin{tabular}[c]{@{}c@{}}Cars 5-shot\end{tabular} \\ \hline
$\checkmark$ &              &              &                                                                                & 70.34 $\pm$ 0.71                                        & 61.16 $\pm$ 0.76                                         \\
             & $\checkmark$ &              &                                                                                & 70.28 $\pm$ 0.70                                        & 61.11 $\pm$ 0.77                                         \\
             &              & $\checkmark$ &                                                                                & 70.92 $\pm$ 0.69                                        & 61.31 $\pm$ 0.76                                         \\
             &              &              & $\checkmark$                                                                   & 70.53 $\pm$ 0.75                                        & 61.53 $\pm$ 0.75                                         \\ \hline
\end{tabular} 
\caption{\textbf{Ablation 2.} Novel task performance with varying sizes of the distractor domain, \ie, \textit{mini}ImageNet-train. Note that this is \textit{different} from distractor batch size $|S_\text{dt}|$.}
\label{tab:absub}
\end{table}
In this section, we introduce a few important ablations that help deconstruct the ConFT and MT-ConFT objectives. In Table \ref{tab:abpn}, we compare our stochastic anchor-positive construction with naive inclusion of all positives for every anchor. We also, ablate the contribution of each type of anchor-negatives: task-specific and distractors and compare that to relative weighting of the two. In Table \ref{tab:absub}, we studied the importance of distractor domain size and found that novel task performance is fairly robust to the size of the distractor domain. This is particularly encouraging since we need not store the entire base data during finetuning. 

\subsection{ConFT for Unsupervised Prior Learning}
\label{exp:ufsl}

\begin{table}
\scriptsize
\centering   
\begin{tabular}{ccccc}
\hline
\multicolumn{2}{c}{Method}                              &          & \multicolumn{2}{c}{5-way \textit{mini}ImageNet} \\ \cline{1-2} \cline{4-5} 
Prior Learning               & Finetuning & BB & 1-shot             & 5-shot             \\ \hline
Sup. MML              & MFT                  & Conv4    & $46.81 _{\pm 0.77}$      & $62.13 _{\pm 0.72}$      \\
Sup. PN         & -                        & Conv4    & $46.56 _{\pm 0.76}$ & $62.29 _{\pm 0.71}$      \\ \hline
-                            & RandInit    & Conv4    & $27.59 _{\pm 0.59}$      & $38.48 _{\pm 0.66}$      \\
BG-MML \cite{hsu2018unsupervised}            & MFT                  & Conv4    & $36.24 _{\pm 0.74}$      & $51.28 _{\pm 0.6}$       \\
BG-PN \cite{hsu2018unsupervised}       & -                        & Conv4    & $36.62 _{\pm 0.70}$      & $50.16 _{\pm 0.7}$       \\
DC-MML   \cite{hsu2018unsupervised}     & MFT                  & Conv4    & $39.90 _{\pm 0.74}$      & $53.97 _{\pm 0.70}$      \\
DC-PN \cite{hsu2018unsupervised}  & -                        & Conv4    & $39.18 _{\pm 0.71}$      & $53.36_{\pm0.70}$        \\
U-MML  \cite{khodadadeh2019unsupervised}                & MFT                  & Conv4    & $39.93$             & $50.73$             \\
LG-MML   \cite{khodadadeh2021unsupervised}         & MFT                  & Conv4    & $40.19 _{\pm 0.58}$      &$54.56 _{\pm 0.55}$      \\
LG-PN  \cite{khodadadeh2021unsupervised}       & -                        & Conv4    & $40.05_{\pm0.60}$        & $52.53 _{\pm 0.51}$      \\ \hline
SimCLR   \cite{chen2020simple}                    & -                        & Conv4    & $41.54 _{\pm 0.61}$     & $56.57 _{\pm 0.59}$     \\
SimCLR                       & ConFT            & Conv4    & $\mathbf{43.45 _{\pm 0.60}}$     & $\mathbf{60.02 _{\pm 0.57}}$     \\ \hline
\end{tabular}
\caption{\textbf{Unsupervised Prior Learning.} The results are averaged over 1000 novel tasks and are presented with $95\%$ confidence intervals. Here, MFT refers to meta-style fintuning \cite{finn2017model}. MML:Maml, PN: ProtoNet, DC: DeepCluster, U: Umtra, LG: Lasium-Gan, BG: BiGAN, Sup.:Supervised, BB: Backbone.}
\label{tab:ufsl}
\end{table}
 In Table \ref{tab:ufsl},  we demonstrate the generality of  contrastive finetuning by evaluating on the unsupervised prior learning benchmark \textit{mini}ImageNet. The key distinction from cross-domain settings is that we do not have labelled base data to learn from. So, we leverage self-supervised contrastive learning \cite{chen2020simple} on   
the unlabelled base data and show that it outperforms state of the art by $1$ to $2$ points. Finetuning the resultant representation with our ConFT objective further improves the accuracy by  $2$to $4$ points. This is particularly significant, as the results come very close to supervised baselines that serve as performance upper bound in this setting \cite{khodadadeh2021unsupervised}. 

\section{Conclusion}
We introduce a novel contrastive finetuning approach to few-shot classification. Specifically, our method leverages distractors to improve generalization by encouraging cluster separation of the novel task samples. 
We show that our method leads to significant performance gains in both cross-domain and unsupervised prior learning setups.

{\small
\bibliographystyle{ieee_fullname}
\bibliography{main}

\begin{thebibliography}{10}\itemsep=-1pt

\bibitem{afrasiyabi2020associative}
Arman Afrasiyabi, Jean-Fran{\c{c}}ois Lalonde, and Christian Gagn{'e}.
\newblock Associative alignment for few-shot image classification.
\newblock In {\em ECCV}, 2020.

\bibitem{antoniou2018data}
Antreas Antoniou, Amos~J. Storkey, and Harrison~A Edwards.
\newblock Data augmentation generative adversarial networks.
\newblock {\em arXiv}, 2017.

\bibitem{NIPS2006BenDA2983}
Shai Ben-David, John Blitzer, Koby Crammer, and Fernando Pereira.
\newblock Analysis of representations for domain adaptation.
\newblock In {\em NeurIPS}. 2007.

\bibitem{bromley1993signature}
Jane Bromley, James Bentz, Leon Bottou, Isabelle Guyon, Yann Lecun, Cliff
  Moore, Eduard Sackinger, and Rookpak Shah.
\newblock Signature verification using a "siamese" time delay neural network.
\newblock {\em International Journal of Pattern Recognition and Artificial
  Intelligence}, 7:25, 1993.

\bibitem{Bukchin2020FinegrainedAC}
Guy Bukchin, Eli Schwartz, Kate Saenko, Ori Shahar, R. Feris, Raja Giryes, and
  Leonid Karlinsky.
\newblock Fine-grained angular contrastive learning with coarse labels.
\newblock {\em arXiv}, 2020.

\bibitem{chen2020simple}
Ting Chen, Simon Kornblith, Mohammad Norouzi, and Geoffrey Hinton.
\newblock A simple framework for contrastive learning of visual
  representations.
\newblock {\em arXiv}, 2020.

\bibitem{chen2019closer}
Wei-Yu Chen, Yen-Cheng Liu, Zsolt Kira, Yu-Chiang~Frank Wang, and Jia-Bin
  Huang.
\newblock A closer look at few-shot classification.
\newblock In {\em ICLR}, 2019.

\bibitem{Chen2020ANM}
Yinbo Chen, Xiaolong Wang, Zhuang Liu, Huijuan Xu, and Trevor Darrell.
\newblock A new meta-baseline for few-shot learning.
\newblock {\em arXiv}, 2020.

\bibitem{Cubuk2018AutoAugmentLA}
Ekin~D. Cubuk, Barret Zoph, Dandelion Man{\'e}, Vijay Vasudevan, and Quoc~V.
  Le.
\newblock Autoaugment: Learning augmentation policies from data.
\newblock {\em arXiv}, 2018.

\bibitem{Dhillon2020A}
Guneet~Singh Dhillon, Pratik Chaudhari, Avinash Ravichandran, and Stefano
  Soatto.
\newblock A baseline for few-shot image classification.
\newblock In {\em ICLR}, 2020.

\bibitem{Doersch2020CrossTransformersSF}
Carl Doersch, Ankush Gupta, and Andrew Zisserman.
\newblock Crosstransformers: Spatially-aware few-shot transfer.
\newblock In {\em NeurIPS}, 2020.

\bibitem{dvornik2019diversity}
Nikita Dvornik, Cordelia Schmid, and Julien Mairal.
\newblock Diversity with cooperation: Ensemble methods for few-shot
  classification.
\newblock In {\em ICCV}, 2019.

\bibitem{finn2017model}
Chelsea Finn, Pieter Abbeel, and Sergey Levine.
\newblock Model-agnostic meta-learning for fast adaptation of deep networks.
\newblock In {\em ICML}, 2017.

\bibitem{Ganin2017}
Yaroslav Ganin, Evgeniya Ustinova, Hana Ajakan, Pascal Germain, Hugo
  Larochelle, François Laviolette, Mario Marchand, and Victor Lempitsky.
\newblock Domain-adversarial training of neural networks.
\newblock {\em CVPR}, 2017.

\bibitem{Ge2017CVPR}
Weifeng Ge and Yizhou Yu.
\newblock Borrowing treasures from the wealthy: Deep transfer learning through
  selective joint fine-tuning.
\newblock In {\em CVPR}, 2017.

\bibitem{gidaris2019boosting}
Spyros Gidaris, Andrei Bursuc, Nikos Komodakis, Patrick P{\'e}rez, and Matthieu
  Cord.
\newblock Boosting few-shot visual learning with self-supervision.
\newblock In {\em CVPR}, 2019.

\bibitem{Gidaris2018DynamicFV}
Spyros Gidaris and Nikos Komodakis.
\newblock Dynamic few-shot visual learning without forgetting.
\newblock {\em CVPR}, 2018.

\bibitem{guo2020broader}
Yunhui Guo, Noel~C Codella, Leonid Karlinsky, James~V Codella, John~R Smith,
  Kate Saenko, Tajana Rosing, and Rogerio Feris.
\newblock A broader study of cross-domain few-shot learning.
\newblock In {\em ECCV}, 2020.

\bibitem{guo2018spottune}
Yunhui Guo, Honghui Shi, Abhishek Kumar, Kristen Grauman, Tajana Rosing, and
  Rogerio Feris.
\newblock Spottune: Transfer learning through adaptive fine-tuning.
\newblock {\em arXiv}, 2018.

\bibitem{pmlr-v9-gutmann10a}
Michael Gutmann and Aapo Hyvärinen.
\newblock Noise-contrastive estimation: A new estimation principle for
  unnormalized statistical models.
\newblock In {\em AISTATS}, 2010.

\bibitem{1640964}
Raia Hadsell, Sumit Chopra, and Yann LeCun.
\newblock Dimensionality reduction by learning an invariant mapping.
\newblock In {\em CVPR}, 2006.

\bibitem{hariharan2016low}
Bharath Hariharan and Ross Girshick.
\newblock Low-shot visual recognition by shrinking and hallucinating features.
\newblock In {\em ICCV}, 2017.

\bibitem{He2020CVPR}
Kaiming He, Haoqi Fan, Yuxin Wu, Saining Xie, and Ross Girshick.
\newblock Momentum contrast for unsupervised visual representation learning.
\newblock In {\em CVPR}, 2020.

\bibitem{He2016DeepRL}
Kaiming He, X. Zhang, Shaoqing Ren, and Jian Sun.
\newblock Deep residual learning for image recognition.
\newblock In {\em CVPR}, 2016.

\bibitem{hjelm2018learning}
R~Devon Hjelm, Alex Fedorov, Samuel Lavoie-Marchildon, Karan Grewal, Phil
  Bachman, Adam Trischler, and Yoshua Bengio.
\newblock Learning deep representations by mutual information estimation and
  maximization.
\newblock In {\em ICLR}, 2019.

\bibitem{hsu2018unsupervised}
Kyle Hsu, Sergey Levine, and Chelsea Finn.
\newblock Unsupervised learning via meta-learning.
\newblock In {\em ICLR}, 2019.

\bibitem{Hnaff2020DataEfficientIR}
Olivier~J. Hénaff, Aravind Srinivas, Jeffrey~De Fauw, Ali Razavi, Carl
  Doersch, S.~M.~Ali Eslami, and Aaron van~den Oord.
\newblock Data-efficient image recognition with contrastive predictive coding.
\newblock {\em arXiv}, 2020.

\bibitem{Kamnitsas2018SemiSupervisedLV}
Konstantinos Kamnitsas, Daniel~C. Castro, Lo{\"i}c~Le Folgoc, Ian Walker,
  Ryutaro Tanno, Daniel Rueckert, Ben Glocker, Antonio Criminisi, and Aditya~V.
  Nori.
\newblock Semi-supervised learning via compact latent space clustering.
\newblock In {\em ICML}, 2018.

\bibitem{khodadadeh2019unsupervised}
Siavash Khodadadeh, Ladislau Boloni, and Mubarak Shah.
\newblock Unsupervised meta-learning for few-shot image classification.
\newblock In {\em NeurIPS}, 2019.

\bibitem{khodadadeh2021unsupervised}
Siavash Khodadadeh, Sharare Zehtabian, Saeed Vahidian, Weijia Wang, Bill Lin,
  and Ladislau Boloni.
\newblock Unsupervised meta-learning through latent-space interpolation in
  generative models.
\newblock In {\em ICLR}, 2021.

\bibitem{NEURIPS2020SupCond89a66c7}
Prannay Khosla, Piotr Teterwak, Chen Wang, Aaron Sarna, Yonglong Tian, Phillip
  Isola, Aaron Maschinot, Ce Liu, and Dilip Krishnan.
\newblock Supervised contrastive learning.
\newblock In {\em NeurIPS}, 2020.

\bibitem{Kingma2015AdamAM}
Diederik~P. Kingma and Jimmy Ba.
\newblock Adam: A method for stochastic optimization.
\newblock {\em CoRR}, 2015.

\bibitem{Koch2015SiameseNN}
Gregory~R. Koch.
\newblock Siamese neural networks for one-shot image recognition.
\newblock In {\em ICML Deep Learning Workshop. Vol. 2.}, 2015.

\bibitem{Kornblith2019DoBI}
Simon Kornblith, Jonathon Shlens, and Quoc~V. Le.
\newblock Do better imagenet models transfer better?
\newblock {\em CVPR}, 2019.

\bibitem{6755945}
Jonathan Krause, Michael Stark, Jia Deng, and Li Fei-Fei.
\newblock 3d object representations for fine-grained categorization.
\newblock In {\em ICCV Workshops}, 2013.

\bibitem{Lake2015HumanlevelCL}
Brenden~M. Lake, Ruslan Salakhutdinov, and Joshua~B. Tenenbaum.
\newblock Human-level concept learning through probabilistic program induction.
\newblock {\em Science}, 350:1332 -- 1338, 2015.

\bibitem{NEURIPS2019bf25356f}
Xinzhe Li, Qianru Sun, Yaoyao Liu, Qin Zhou, Shibao Zheng, Tat-Seng Chua, and
  Bernt Schiele.
\newblock Learning to self-train for semi-supervised few-shot classification.
\newblock In {\em NeurIPS}, 2019.

\bibitem{liang-etal-2018-distractor}
Chen Liang, Xiao Yang, Neisarg Dave, Drew Wham, Bart Pursel, and C.~Lee Giles.
\newblock Distractor generation for multiple choice questions using learning to
  rank.
\newblock In {\em Proceedings of the Thirteenth Workshop on Innovative Use of
  {NLP} for Building Educational Applications}, 2018.

\bibitem{Majumder2021RevisitingCL}
Orchid Majumder, Avinash Ravichandran, Subhransu Maji, M. Polito, Rahul
  Bhotika, and Stefano Soatto.
\newblock Revisiting contrastive learning for few-shot classification.
\newblock {\em arXiv}, 2021.

\bibitem{munkhdalai2017meta}
Munkhdalai, Tsendsuren, Yu, and Hong.
\newblock Meta networks.
\newblock In {\em ICML}, 2017.

\bibitem{nichol2018reptile}
Alex Nichol and John Schulman.
\newblock Reptile: A scalable metalearning algorithm.
\newblock {\em arXiv}, 2018.

\bibitem{oh2021boil}
Jaehoon Oh, Hyungjun Yoo, ChangHwan Kim, and Se-Young Yun.
\newblock {\{}BOIL{\}}: Towards representation change for few-shot learning.
\newblock In {\em ICLR}, 2021.

\bibitem{oliver2018realistic}
Avital Oliver, Augustus Odena, Colin Raffel, Ekin~D Cubuk, and Ian~J
  Goodfellow.
\newblock Realistic evaluation of deep semi-supervised learning algorithms.
\newblock {\em arXiv}, 2018.

\bibitem{ouali2020spatial}
Yassine Ouali, C{\'e}line Hudelot, and Myriam Tami.
\newblock Spatial contrastive learning for few-shot classification.
\newblock {\em arXiv}, 2020.

\bibitem{5288526}
Sinno~Jialin Pan and Qiang Yang.
\newblock A survey on transfer learning.
\newblock {\em IEEE Transactions on Knowledge and Data Engineering}, 2010.

\bibitem{phoo2021selftraining}
Cheng~Perng Phoo and Bharath Hariharan.
\newblock Self-training for few-shot transfer across extreme task differences.
\newblock In {\em ICLR}, 2021.

\bibitem{PosseggerCVPR}
Horst Possegger, Thomas Mauthner, and Horst Bischof.
\newblock In defense of color-based model-free tracking.
\newblock In {\em CVPR}, June 2015.

\bibitem{46678}
Hang Qi, David Lowe, and Matthew Brown.
\newblock Low-shot learning with imprinted weights.
\newblock In {\em CVPR}, 2018.

\bibitem{Raghu2020Rapid}
Aniruddh Raghu, Maithra Raghu, Samy Bengio, and Oriol Vinyals.
\newblock Rapid learning or feature reuse? towards understanding the
  effectiveness of maml.
\newblock In {\em ICLR}, 2020.

\bibitem{Ravi2017OptimizationAA}
Sachin Ravi and Hugo Larochelle.
\newblock Optimization as a model for few-shot learning.
\newblock In {\em ICLR}, 2017.

\bibitem{ravi2017optimization}
Sachin Ravi and Hugo Larochelle.
\newblock Optimization as a model for few-shot learning.
\newblock In {\em ICLR}, 2017.

\bibitem{ren2018metalearning}
Mengye Ren, Sachin Ravi, Eleni Triantafillou, Jake Snell, Kevin Swersky,
  Josh~B. Tenenbaum, Hugo Larochelle, and Richard~S. Zemel.
\newblock Meta-learning for semi-supervised few-shot classification.
\newblock In {\em ICLR}, 2018.

\bibitem{robinson2021contrastive}
Joshua~David Robinson, Ching-Yao Chuang, Suvrit Sra, and Stefanie Jegelka.
\newblock Contrastive learning with hard negative samples.
\newblock In {\em ICLR}, 2021.

\bibitem{ILSVRC15}
Olga Russakovsky, Jia Deng, Hao Su, Jonathan Krause, Sanjeev Satheesh, Sean Ma,
  Zhiheng Huang, Andrej Karpathy, Aditya Khosla, Michael Bernstein,
  Alexander~C. Berg, and Li Fei-Fei.
\newblock {ImageNet Large Scale Visual Recognition Challenge}.
\newblock {\em IJCV}, 2015.

\bibitem{rusu2019meta}
Andrei~A. Rusu, Dushyant Rao, Jakub Sygnowski, Oriol Vinyals, Razvan Pascanu,
  Simon Osindero, and Raia Hadsell.
\newblock Meta-learning with latent embedding optimization.
\newblock In {\em ICLR}, 2019.

\bibitem{Ryu2020MetaPerturbTR}
Jeongun Ryu, Jaewoong Shin, H.~B. Lee, and Sung~Ju Hwang.
\newblock Metaperturb: Transferable regularizer for heterogeneous tasks and
  architectures.
\newblock {\em arXiv}, 2020.

\bibitem{Saikia2020OptimizedGF}
Tonmoy Saikia, T. Brox, and C. Schmid.
\newblock Optimized generic feature learning for few-shot classification across
  domains.
\newblock {\em arXiv}, 2020.

\bibitem{pmlr-v2-salakhutdinov07a}
Ruslan Salakhutdinov and Geoff Hinton.
\newblock Learning a nonlinear embedding by preserving class neighbourhood
  structure.
\newblock In {\em AISTATS}, 2007.

\bibitem{garcia2018fewshot}
Victor~Garcia Satorras and Joan~Bruna Estrach.
\newblock Few-shot learning with graph neural networks.
\newblock In {\em ICLR}, 2018.

\bibitem{pmlr-v97-saunshi19a}
Nikunj Saunshi, Orestis Plevrakis, Sanjeev Arora, Mikhail Khodak, and
  Hrishikesh Khandeparkar.
\newblock A theoretical analysis of contrastive unsupervised representation
  learning.
\newblock In {\em ICML}, 2019.

\bibitem{Snell2017PrototypicalNF}
Jake Snell, Kevin Swersky, and Richard~S. Zemel.
\newblock Prototypical networks for few-shot learning.
\newblock In {\em NeurIPS}, 2017.

\bibitem{NIPS20166b180037}
Kihyuk Sohn.
\newblock Improved deep metric learning with multi-class n-pair loss objective.
\newblock In {\em NeurIPS}, 2016.

\bibitem{Su2020When}
Jong-Chyi Su, Subhransu Maji, and Bharath Hariharan.
\newblock When does self-supervision improve few-shot learning?
\newblock In {\em ECCV}, 2020.

\bibitem{sung2018learning}
Flood Sung, Yongxin Yang, Li Zhang, Tao Xiang, Philip~H.S. Torr, and Timothy~M.
  Hospedales.
\newblock Learning to compare: Relation network for few-shot learning.
\newblock In {\em CVPR}, 2018.

\bibitem{tian2019contrastive}
Yonglong Tian, Dilip Krishnan, and Phillip Isola.
\newblock Contrastive multiview coding.
\newblock {\em arXiv}, 2019.

\bibitem{tian2020rethink}
Yonglong Tian, Yue Wang, Dilip Krishnan, Joshua~B Tenenbaum, and Phillip Isola.
\newblock Rethinking few-shot image classification: A good embedding is all you
  need?
\newblock {\em arXiv}, 2020.

\bibitem{triantafillou2019metadataset}
Eleni Triantafillou, Tyler Zhu, Vincent Dumoulin, Pascal Lamblin, Utku Evci,
  Kelvin Xu, Ross Goroshin, Carles Gelada, Kevin Swersky, Pierre-Antoine
  Manzagol, and Hugo Larochelle.
\newblock Meta-dataset: A dataset of datasets for learning to learn from few
  examples.
\newblock In {\em ICLR}, 2020.

\bibitem{crossdomainfewshot}
Hung-Yu Tseng, Hsin-Ying Lee, Jia-Bin Huang, and Ming-Hsuan Yang.
\newblock Cross-domain few-shot classification via learned feature-wise
  transformation.
\newblock In {\em ICLR}, 2020.

\bibitem{Tseng2020Cross-Domain}
Hung-Yu Tseng, Hsin-Ying Lee, Jia-Bin Huang, and Ming-Hsuan Yang.
\newblock Cross-domain few-shot classification via learned feature-wise
  transformation.
\newblock In {\em ICLR}, 2020.

\bibitem{Oord2018RepresentationLW}
Aaron van~den Oord, Yazhe Li, and Oriol Vinyals.
\newblock Representation learning with contrastive predictive coding.
\newblock {\em arXiv}, 2018.

\bibitem{JMLR:v9:vandermaaten08a}
Laurens van~der Maaten and Geoffrey Hinton.
\newblock Visualizing data using t-sne.
\newblock {\em JMLR}, 9, 2008.

\bibitem{van2018inaturalist}
Grant Van~Horn, Oisin Mac~Aodha, Yang Song, Yin Cui, Chen Sun, Alex Shepard,
  Hartwig Adam, Pietro Perona, and Serge Belongie.
\newblock The inaturalist species classification and detection dataset.
\newblock In {\em CVPR}, 2016.

\bibitem{NIPS2016Match6385}
Oriol Vinyals, Charles Blundell, Timothy Lillicrap, koray kavukcuoglu, and Daan
  Wierstra.
\newblock Matching networks for one shot learning.
\newblock In {\em NIPS}. 2016.

\bibitem{WangChest2017}
Xiaosong Wang, Yifan Peng, Le Lu, Zhiyong Lu, Mohammadhadi Bagheri, and
  Ronald~M. Summers.
\newblock Chestx-ray8: Hospital-scale chest x-ray database and benchmarks on
  weakly-supervised classification and localization of common thorax diseases.
\newblock {\em 2017 IEEE Conference on Computer Vision and Pattern Recognition
  (CVPR)}, 2017.

\bibitem{wang2018low}
Yu-Xiong Wang, Ross Girshick, Martial Hebert, and Bharath Hariharan.
\newblock Low-shot learning from imaginary data.
\newblock In {\em CVPR}, 2018.

\bibitem{Wang-2016-26050}
Yu-Xiong Wang and Martial Hebert.
\newblock Learning from small sample sets by combining unsupervised
  meta-training with cnns.
\newblock In {\em NeurIPS}, 2016.

\bibitem{WelinderEtal2010}
Peter Welinder, Steve Branson, Takeshi Mita, Catherine Wah, Florian Schroff,
  Serge Belongie, and Pietro Perona.
\newblock {Caltech-UCSD Birds 200}.
\newblock Technical Report CNS-TR-2010-001, California Institute of Technology,
  2010.

\bibitem{wu2018improving}
Zhirong Wu, Alexei~A Efros, and Stella Yu.
\newblock Improving generalization via scalable neighborhood component
  analysis.
\newblock In {\em ECCV}, 2018.

\bibitem{wu2018unsupervised}
Zhirong Wu, Yuanjun Xiong, X~Yu Stella, and Dahua Lin.
\newblock Unsupervised feature learning via non-parametric instance
  discrimination.
\newblock In {\em CVPR}, 2018.

\bibitem{NIPS2014375c7134}
Jason Yosinski, Jeff Clune, Yoshua Bengio, and Hod Lipson.
\newblock How transferable are features in deep neural networks?
\newblock In {\em NeurIPS}, 2014.

\bibitem{7968387}
Bolei Zhou, Agata Lapedriza, Aditya Khosla, Aude Oliva, , and Antonio Torralba.
\newblock Places: A 10 million image database for scene recognition.
\newblock {\em TPAMI}, 40(6):1452--1464, 2018.

\bibitem{Zhu18ECCV}
Zheng Zhu, Qiang Wang, Li Bo, Wei Wu, Junjie Yan, and Weiming Hu.
\newblock Distractor-aware siamese networks for visual object tracking.
\newblock In {\em ECCV}, 2018.

\end{thebibliography}
} 

\cleardoublepage
\appendix
\section{Overview of ConFT}
 \begin{algorithm*}[t]
  \caption{Distractor-Aware Contrastive Finetuning}
  \begin{algorithmic}[1]
  \Require{Distractor Dataset ($D$), Prior Model ($\mathcal{M}_{\theta_0}$), few-shot task ($\tau$), Number of Finetuning Epochs ($J_\text{ft}$), Augmentation Function ($\mathcal{A}$), Temperature Coefficient ($\gamma$), Learning Rate ($\eta$)}
 
 \Ensure{Finetuned Model Parameters ($\theta_\tau$)}
\State shuffle $D$
\For{j  $\gets 1~ \text{to} ~J_\text{ft}$ }
  \State From $D$, randomly sample a fixed size batch $S_\text{dt}$ without replacement
  \State Using $\mathcal{A}$ augment each support sample $x_i, ~\forall i\in I_\text{supp}$
  \State For each augmented support sample, define i) anchor-positive index set $P(i)$; ii)  anchor-negative index set $N(i)$ specific to $\tau$; and iii) distractor index set $I_\text{dt}$ 
 \State For all samples, compute $z_i = \frac{h_i}{||h_i||_2}$, where $h=\mathcal{M}_\theta(\mathcal{A}(x_i)),~ \forall  i\in I_\text{supp}$  and  $h=\mathcal{M}_\theta(x_i),~ \forall  i\in I_\text{dt}$
   \State Evaluate $\mathcal{L}_\text{conft}(\theta) $ using the quantities computed in previous steps
  \State Update model parameters $\theta \gets \theta - \eta \nabla \mathcal{L}_\text{conft}(\theta)$
  \If{$j=|D|$}
  \State shuffle $D$
  \EndIf
 \EndFor 
  \end{algorithmic}
  \label{alg:conft}
  \end{algorithm*} 
  
Algorithm \ref{alg:conft} provides an overview of our distractor-aware contrastive finetuning approach ConFT.

\section{Additional Experimental Details}

\subsection{Data Domains}

\begin{table}[htbp]
\centering
\resizebox{\columnwidth}{!}{ 
\begin{tabular}{cccccc}
\hline
Problem Setup\\(Prior Learning)                                                         & Domain                                                      & Dataset      & \multicolumn{3}{c}{$\#$ categories per split} \\
                                                                      &                                                             &              & train            & val            & test           \\ \hline
 & Base                                                        & \textit{mini}ImageNet & 64               & 16             & 20             \\
                                                                      &                       Novel                                 & CUB          & 100              & 50             & 50             \\
     \begin{tabular}[c]{@{}c@{}}Cross-Domain\end{tabular}
     &        Novel                                                     & Cars         & 98               & 49             & 49             \\
                                                                      &                          Novel                                   & Places       & 183              & 91             & 91             \\
                                                                      &                            Novel                                 & Plantae      & 100              & 50             & 50             \\\hline
\begin{tabular}[c]{@{}c@{}}Unsupervised\end{tabular} & \begin{tabular}[c]{@{}c@{}}Base \\ and\\ Novel\end{tabular} & \textit{mini}ImageNet & 64               & 16             & 20             \\\hline
\multicolumn{1}{l}{}                                                  & \multicolumn{1}{l}{}                                        & \multicolumn{1}{l}{} & \multicolumn{1}{l}{} & \multicolumn{1}{l}{} & \multicolumn{1}{l}{}
\end{tabular}}
\caption{\textbf{Dataset statistics for both cross-domain \cite{crossdomainfewshot} and unsupervised prior learning settings \cite{hsu2018unsupervised}.} Each dataset is split into \textit{train}, \textit{val}, and \textit{test} categories. For the cross-domain setup, the \textit{train} split of \textit{mini}ImageNet is always used as the base domain whereas the \textit{test} splits of other datasets are used as the novel domain on which few-shot evaluation is performed. For the unsupervised prior learning setup, \textit{train} split of \textit{mini}ImageNet is stripped off its labels to emulate an unlabelled base domain, whereas the \textit{test} split is used as the novel domain. In both setups, \textit{val} splits are used to cross-validate hyperparameters specific to the associated novel domain.}
\label{tab:datastat}
\end{table}

\begin{table}[t]
\centering
\includegraphics[width=0.9\linewidth]{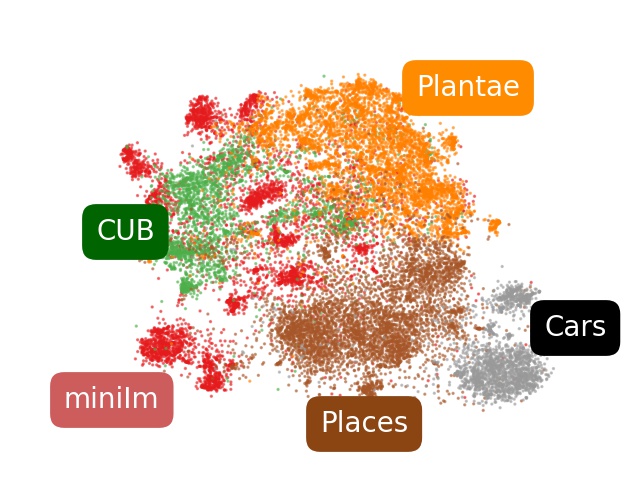}
\bigskip
\begin{tabular}{cccc}
\hline
\begin{tabular}[c]{@{}c@{}}Base\\ Domain\end{tabular} & \begin{tabular}[c]{@{}c@{}}Novel\\ Domain\end{tabular} & \begin{tabular}[c]{@{}c@{}}Proximity\\ Rank\end{tabular} & PAD  \\ \hline
                    & Places                                                 & 1                                                        & 1.06 \\
\textit{mini}ImageNet                                                & CUB                                                    & 2                                                        & 1.57 \\
                                                         & Plantae                                                & 3                                                        & 1.76 \\
                                                      & Cars                                                   & 4                                                        & 1.86 \\ \hline
\end{tabular}
\caption{\textbf{Qualitative and quantitative visualization of the base and novel domains in the cross-domain benchmark \cite{crossdomainfewshot}.} We use t-SNE to visualize the base and novel domains in our cross-domain benchmark. The domain names are presented in boxes with colors that match the corresponding domains in the scatter plot. Here,``miniIm" refers to the \textit{mini}ImageNet domain. We also compute the \textit{Proxy $\mathcal{A}$-distance} (PAD) \cite{NIPS2006BenDA2983,Ganin2017} between the base domain and a novel domain as a measure of their relatedness in the representation space. Smaller the PAD value, closer is the novel domain to the base and hence, more related. The PAD values are also used to rank the novel domains according to their proximity to the base domain with the closest domain ranked the highest. }
\label{tab:PAD}
\end{table}

In the main paper, we evaluated our finetuning method on various datasets that serve as base or novel domains in cross-domain as well as unsupervised prior learning settings. Here, in  Table \ref{tab:datastat}, we summarize the statistics of these datasets along with their specific use as base or novel domain. Additionally, in Table \ref{tab:PAD}, we visualize these domains, both qualitatively and quantitatively, to provide a reference to their relative proximity in the representation space. This proximity provides a rough estimate of how related two domains are and consequently, the degree of knowledge transfer across domains for cross-domain few-shot classification.

For the qualitative visualization in Table \ref{tab:PAD}, we use t-SNE \cite{JMLR:v9:vandermaaten08a} to embed features of randomly sampled datapoints from each domain onto a $2$-dimensional space. These features are obtained from the pretrained ResNet10 model (see \S \ref{sec:pl} for training details) and are used for our cross-domain experiments. For quantitative visualization, we compute  \textit{Proxy $\mathcal{A}$-distance} \cite{NIPS2006BenDA2983,Ganin2017}, or $\text{PAD}$,  between the base domain (here, \textit{mini}ImageNet) and a novel domain as a measure of their closeness in the representation space.  To compute $\text{PAD}$, we train a binary classifier over the same ResNet10 model used for t-SNE but with frozen embedding weights. The classifier distinguishes between randomly drawn samples of the base and novel domains. Denoting $\epsilon$ as the generalization error of this classifier, the $\text{PAD} \in [0,2]$ is calculated as $2(1-2\epsilon)$. Thus, a lower $\text{PAD}$ value implies higher generalization error which, in turn, signifies that the base and novel domains are too similar to be distinguished well enough. Finally, the PAD values are used to rank each novel domain, such that the highest rank is assigned to the one  closest to the base domain \ie, \textit{mini}ImageNet. These ranks correlate well with the t-SNE visualization as well. For instance, CUB and Places, which are ranked higher than Cars and Plantae, are also closer to \textit{mini}ImageNet in the t-SNE plot. 

\subsection{Prior Learning}
\label{sec:pl}

As described in the main paper, we use a ResNet10 model \cite{He2016DeepRL} as our prior embedding for cross-domain few-shot classification. To avoid specialized hyperparameter tuning while training the prior model, we simply use the pretrained weights\footnote{\url{https://github.com/hytseng0509/CrossDomainFewShot}} made available by \cite{crossdomainfewshot}. This model was originally trained on all $64$ categories of the \textit{mini}ImageNet \textit{train} split. 

For the unsupervised prior learning, we train a modified four-layer convolution neural network (CNN), using the recently proposed self-supervised contrastive learning objective \cite{chen2020simple}. As proposed in \cite{chen2020simple}, we use a $128$-dimensional linear projection head on top of the CNN for better generalizability of learnt representations. We train the model with a batch size of $512$, temperature coefficient $0.1$, and the same augmentation scheme introduced in \cite{chen2020simple}. Further, we use ADAM optimizer with initial learning rate of $1e-3$, and a weight decay of $1e-5$.

\subsection{Hyperparameter Details}
\begin{table*}[htbp]
\centering
\resizebox{1.8\columnwidth}{!}{ 
\begin{tabular}{cccccccccc}
\hline
                     &                            & \multicolumn{2}{c}{CUB}                     & \multicolumn{2}{c}{Cars}                    & \multicolumn{2}{c}{Places}                  & \multicolumn{2}{c}{Plantae}                 \\
Hyperparameter       & Range                      & 1-shot               & 5-shot               & 1-shot               & 5-shot               & 1-shot               & 5-shot               & 1-shot               & 5-shot               \\ \hline
learning rate        & \{5e-4, 5e-3\}             & 5e-3                 & 5e-3                 & 5e-3                 & 5e-3                 & 5e-4                 & 5e-4                 & 5e-3                 & 5e-3                 \\
temperature, $\gamma$            & \{0.05, 0.1, 0.5\}         & 0.1                  & 0.1                  & 0.05                 & 0.05                 & 0.1                  & 0.05                 & 0.1                  & 0.1                  \\
distractor batch size, $|S_\text{dt}|$  & \{64, 128\}                & 64                   & 128                  & 128                  & 128                  & 64                   & 64                   & 128                  & 128                  \\
early stopping epoch & \{50, 100, 200, 300, 400\} & 100                  & 100                  & 400                  & 300                  & 200                  & 50                   & 100                  & 100                  \\ \hline
\multicolumn{1}{l}{} & \multicolumn{1}{l}{}       & \multicolumn{1}{l}{} & \multicolumn{1}{l}{} & \multicolumn{1}{l}{} & \multicolumn{1}{l}{} & \multicolumn{1}{l}{} & \multicolumn{1}{l}{} & \multicolumn{1}{l}{} & \multicolumn{1}{l}{}
\end{tabular}}
\caption{\textbf{Hyperparameter details for ConFT with cross-domain prior learning.} This table summarizes the range of various hyperparameters used for finetuning. Additionally, we report the cross-validated values used for the cross-domain prior learning setup. The input image resolution used in this setup is $224\times224$.}
\label{tab:hyperparamC}
\end{table*}

\begin{table}[t]
\centering
\resizebox{\columnwidth}{!}{ 
\begin{tabular}{cccc}
\hline
                     &                                  & \multicolumn{2}{c}{\textit{mini}ImageNet} \\
Hyperparameter       & Range                            & 1-shot          & 5-shot         \\ \hline
learning rate        & \{5e-4, 5e-3\}                   & 5e-4            & 5e-4           \\
temperature, $\gamma$           & \{0.05, 0.1, 0.5\}               & 0.05            & 0.05           \\
distractor batch size, $|S_\text{dt}|$  & \{64, 128\}                      & 64              & 64             \\
early-stopping epoch & \{50, 100, 200, 300, 400,  500\} & 400             & 400            \\ \hline
\multicolumn{1}{l}{} & \multicolumn{1}{l}{}             & \multicolumn{1}{l}{} & \multicolumn{1}{l}{}
\end{tabular} }
\caption{\textbf{Hyperparameter details for ConFT with unsupervised prior learning.} This table summarizes the range of various hyperparameters used for finetuning. Additionally, we report the cross-validated values used for the unsupervised prior learning setup. The input image resolution used in this setup is $84\times84$.}
\label{tab:hyperparamU}
\end{table}
Our proposed contrastive finetuning involves a few hyperparameters such as  temperature, learning rate, early-stopping criteria, distractor batch size, and data augmentation scheme. For early-stopping criteria, we set a predetermined range of epochs up to which the pretrained embedding model is finetuned. Here, one finetuning epoch refers to one pass through all the samples of the few-shot task (exclusive of distractors). The range of these epochs along with other hyperparameters are summarized in Table \ref{tab:hyperparamC} and Table \ref{tab:hyperparamU}. Additionally, we also show the final hyperparameter values used for finetuning in the cross-domain and unsupervised prior learning settings (the corresponding experiments were reported in the main paper). 
  
\section{Additional Ablations }

\begin{table}[htbp]
\centering
\resizebox{.8\columnwidth}{!}{ 
\begin{tabular}{ccc}
\hline
Similarity-Pair Construction& \multicolumn{2}{c}{Cars}                    \\
                                                                                        & 1-shot               & 5-shot               \\ \hline
Standard                                                                                & 37.09 $\pm$ 0.76        & 60.72 $\pm$ 0.74        \\
Assymetric (ours)                                                                       &\textbf{ 39.11 $\pm$ 0.77 }       & \textbf{61.53 $\pm$ 0.75 }       \\ \hline
\multicolumn{1}{l}{}                                                                    & \multicolumn{1}{l}{} & \multicolumn{1}{l}{}
\end{tabular}}
\caption{\textbf{Comparing our proposed assymetric construction of similarity pairs against standard construction:}. Results are shown for both $1$-shot and $5$-shot tasks sampled from the Cars domain with \textit{mini}Imagenet as the base domain. These results are averaged over 600 random novel tasks and are reported with  ($\pm$) $95\%$ confidence intervals. Despite using more supervision in the form of distractor labels, the standard pair construction under-performs our (distractor) label-agnostic asymmetric pair construction.}
\label{tab:assym}
\end{table}

\begin{figure}[h]
\centering 
         \includegraphics[width=\linewidth]{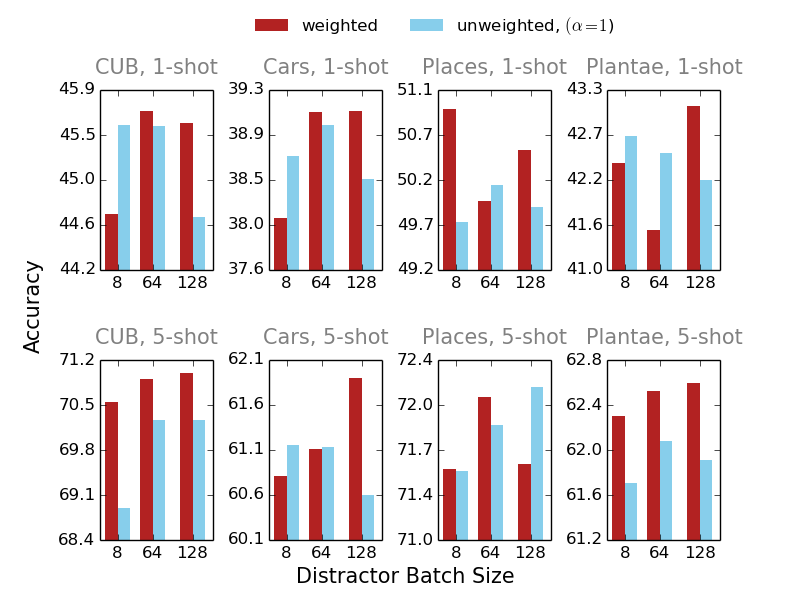}
\caption{\textbf{Comparing the effect of distractor batch size, $|S_\text{dt}|$, on the weighted and unweighted versions of $\mathcal{L}_\text{conft}$.} The \textit{red} and \textit{blue} bars represent weighted and unweighted versions of $\mathcal{L}_\text{conft}$, respectively, where  $\alpha$ represents the parameter used to relatively weigh task-specific and task-exclusive (distractors) anchor-negative terms. For each novel domain and shot setting per domain, we compare the performance of two versions in terms of the classification accuracy of unseen samples given a novel task at various distractor batch sizes. These accuracies, as in all other cross-domain experiments, are averaged over 600 randomly chosen novel tasks.}
\label{fig:rwn}
\end{figure}

In this section we elucidate the importance of two modifications introduced to the standard contrastive loss, namely, asymmetric construction of similarity pairs and relative weighting of anchor-negative terms.

\subsection{Asymmetric Construction of Similarity Pairs}

Our proposed finetuning approach is a general contrastive learning framework for incorporating additional \textit{unlabelled} data in the form of distractors. While construction of positive distractor pairs (that share the same class) is difficult in the absence of distractor labels, constructing anchor-negatives, with anchors being task-specific samples, is much easier following the non-overlapping assumption of task and distractor categories. This results in an asymmetric construction of similarity pairs where distractors, unlike task-specific samples, can meaningfully participate only as anchor-negatives. In fact, this asymmetry is critical in the  unsupervised prior learning setup, where distractors are sampled from an unlabelled base domain. In the case of cross-domain prior learning, however, we have a labelled base data as a source for distractors. To motivate our asymmetric pair construction in this case,  we compare it to a standard construction that allows distractors to additionally participate as anchor-positives.  To form such an anchor-positive, a distractor is paired with another distractor sharing the same class. Here, anchor-negatives with respect to a distractor include all the datapoints that do not share the class with it. This includes samples from both the novel few-shot task and other distractors. Overall, the resulting form of the contrastive loss can be viewed as applying  supervised contrastive objective \cite{NEURIPS2020SupCond89a66c7} (without augmentation-based positives) to the union set of task samples and distractors within a training batch. In Table \ref{tab:assym}, we evaluate these two types of pair constructions on the cross-domain setting, \textit{mini}ImageNet $\to$ Cars. Interestingly enough, our formulation of the contrastive loss with asymmetric pair construction yields superior performance despite using less supervision than the supervised contrastive loss.

\subsection{Importance of Weighted Negatives  }

Another important component of our loss is the relative weighting parameter $\alpha$ that balances the effect of task-specific and task-exclusive (distractor based) anchor-negative terms. To validate the utility of such a weighting scheme, we compare the weighted version of $\mathcal{L}_\text{conft}$  to its unweighted version \ie, $\alpha=1$. Following the results for various novel domains and shot settings in Figure \ref{fig:rwn}, we make the following observations. The weighted loss (\textit{red} bars) performance  improves with larger distractor batch sizes in most cases ($5$ out of $8$). The improvement is more pronounced for domains like Cars and Plantae that are farther away from the base dataset - \textit{mini}ImageNet (see Table \ref{tab:PAD}). For closer domains like CUB or Places, we sometimes notice a sweet spot at batch size $= 64$. In contrast, the unweighted version (\textit{blue} bars) experiences a performance drop with increasing batch sizes, when the novel domains are farther from the base domain. In other cases, the trends are inconclusive. The most important observation, however, comes from comparing the two versions of the loss. Specifically, the weighted version not only outperforms the unweighted loss at higher batch sizes but also results in the best performance in almost every setting. The only exception is Places, 5-shot where the unweighted loss yields the best performance. A possible explanation is as follows: due to the similarity of Places (novel domain) and \textit{mini}ImageNet (base domain) in the embedding space (see Table \ref{tab:PAD}), distractor samples from Places may serve as hard negatives that are important for effective contrastive learning \cite{robinson2021contrastive}. Thus, down-weighting their contribution at higher batch sizes would degrade the final performance.

\subsection{Data Augmentation}

\begin{table}[htbp]
\centering
\resizebox{.88\columnwidth}{!}{ 
\begin{tabular}{cccc}
\hline
\multicolumn{2}{c}{Augmentation}               & CUB           & Cars          \\
Task Samples & \multicolumn{1}{c}{Distractors} & 5-shot        & 5-shot        \\ \hline
      -       &    -     &       69.90 $\pm$ 0.75        & 58.64 $\pm$ 0.88 \\
$\checkmark$ &      -     & 70.53 $\pm$ 0.75 & 61.53 $\pm$ 0.75 \\ \hline
             &                                  &               &              
\end{tabular}}
\caption{In this ablation we compare the few-shot performance when a prior embedding is finetuned (using ConFT) with or without augmentation to task-specific samples. Note that, we never use augmentation for distractors in our experiments.}
\label{tab:noaug}
\end{table}
Yet another important component of our contrastive finetuning objective is the data augmentation function $\mathcal{A}$. To avoid extensive tuning of large hyperparameter space associated with $\mathcal{A}$, we adopt a fixed augmentation strategy introduced in \cite{chen2019closer}. In Table \ref{tab:noaug}, we show the benefit of using this strategy to augment samples specific to the novel task.  Following preliminary investigations, we found that augmenting distractors did not make much difference. Hence, we never apply data augmentation to distractors in our experiments. 

\subsection{Loss Type}
\begin{table}[t]
\centering
\scriptsize 
\begin{tabular}{cccc}
\hline
\multicolumn{2}{c}{Method}                & CUB           & Cars          \\
Prior Learning & Task Specific Finetuning & 5-shot        & 5-shot        \\ \hline
CE Training    & MT-ceFT ($\beta=1$)          & 71.35 $\pm$ 0.70 & 58.97 $\pm$ 0.76 \\
CE Training    & MT-ceFT ($\beta=10$)         & 74.32 $\pm$ 0.69 & 60.01 $\pm$ 0.74 \\
CE Training    & MT-ConFT ($\beta=1$)         & 71.65 $\pm$ 0.74 & 61.25 $\pm$ 0.70 \\
CE Training    & MT-ConFT ($\beta=10$)        & 74.45 $\pm$ 0.71 & 62.54 $\pm$ 0.72 \\ \hline
\end{tabular} 
\caption{\textbf{Ablation.} Cross-entropy/contrastive finetuning with a multi-task (MT) cross entropy objective. Here, all cross entropy objectives are based on cosine classifier with a multiplying factor, $\beta$}
\label{tab:abmt} 
\end{table}

In Table \ref{tab:abmt}, we compare contrastive and cross-entropy finetuning in conjunction with the auxiliary cross-entropy objective (MT). While the two objectives yield similar performance for the CUB case, contrastive finetuning outperforms cross-entropy loss based finetuning in Cars. These results show that the contrastive loss could be a better choice for few-shot classification.

\section{ConFT as a General Finetuning Approach}

\begin{table*}[htbp] 
\centering 
\resizebox{1.8\columnwidth}{!}{ 
\begin{tabular}{ccccccc}
\hline
               &                          &           &                &                &                &                \\

\multicolumn{2}{c}{Method}                &           & \multicolumn{4}{c}{5-shot}                                        \\ \cline{4-7} 
Prior Learning & Task Specific Finetuning & Backbone  & CUB            & Cars           & Places         & Plantae        \\ \hline
ProtoNet \cite{Snell2017PrototypicalNF}    & -                        							&ResNet10  &  58.80 $\pm$ 0.77 & 44.07 $\pm$ 0.69 & 71.03 $\pm$ 0.72& 51.33 $\pm$  0.72\\
ProtoNet \cite{Snell2017PrototypicalNF}    &ConFT (ours)                           &ResNet10  & \textbf{66.63 $\pm$ 0.69} &\textbf{ 59.27 $\pm$ 0.73 }& \textbf{72.05 $\pm$ 0.71} & \textbf{58.83 $\pm$ 0.76}  \\
\hline
ProtoNet + Rot.  \cite{Su2020When} & -                        &ResNet10  & 58.68 $\pm$ 0.75 & 46.48 $\pm$ 0.71 &  71.20 $\pm$ 0.75 & 51.93 $\pm$ 0.67 \\
ProtoNet + Rot.  \cite{Su2020When}   & ConFT (ours)                    &ResNet10  & \textbf{ 66.75 $\pm$ 0.71} & \textbf{61.67 $\pm$ 0.75} &\textbf{ 73.91 $\pm$ 0.70 }& \textbf{60.38 $\pm$ 0.75 }\\
\hline
CE Training    & -                        & ResNet10  & 62.80 $\pm$ 0.76  & 51.41 $\pm$ 0.72  & 70.71 $\pm$ 0.68  & 55.54 $\pm$ 0.69  \\
CE Training    & ConFT (ours)             & ResNet10  & \textbf{70.53 $\pm$ 0.75 } & \textbf{61.53 $\pm$ 0.75 } & \textbf{72.09 $\pm$ 0.68 } & \textbf{62.54 $\pm$ 0.76 } \\
\hline 
\end{tabular} }
\caption{\textbf{Combining ConFT with different pretraining schemes for cross-domain prior learning.} We present the results for $5$-way $5$-shot tasks averaged over 600 such tasks with   ($\pm$) $95\%$ confidence intervals. The highlighted numbers demonstrate that ConFT consistently improves the few-shot performance of prior embeddings across data domains.}
\label{tab:comple}
\end{table*}

\begin{table*}[t]
\centering
\resizebox{1.8\columnwidth}{!}{ 
\begin{tabular}{ccccccc}
\hline
\multicolumn{2}{c}{Method}                &          & \multicolumn{4}{c}{1-shot}                            \\
Prior Learning & Task Specific Finetuning & Backbone & CUB         & Cars        & Places      & Plantae     \\ \hline
SCL     \cite{ouali2020spatial}           & -                        & ResNet12 & 50.09 $\pm$ 0.7 & 34.93 $\pm$ 0.6 & \textbf{60.32 $\pm$ 0.8 }& 40.23 $\pm$ 0.6 \\
CE Training    & -                        & ResNet12 &    50.00 $\pm$ 0.77         &      34.88 $\pm$ 0.64       &  55.62 $\pm$ 0.91           &       38.47 $\pm$ 0.72      \\
CE Training    & ConFT (ours)             & ResNet12 &       \textbf{52.01 $\pm$ 0.82 }     &      \textbf{  39.54 $\pm$ 0.68}     &     56.66 $\pm$ 0.85        &     \textbf{ 40.90 $\pm$ 0.73 }      \\\hline\hline
               &                          &          &             &             &             &             \\ 
\multicolumn{2}{c}{Method}                &          & \multicolumn{4}{c}{5-shot}                            \\
Prior Learning & Task Specific Finetuning & Backbone & CUB         & Cars        & Places      & Plantae     \\ \hline
SCL      \cite{ouali2020spatial}          & -                        & ResNet12 & 68.81 $\pm$ 0.6 & 52.22 $\pm$ 0.7 & \textbf{76.51 $\pm$ 0.6} & \textbf{59.91 $\pm$ 0.6} \\
CE Training    & -                        & ResNet12  &   69.75 $\pm$ 0.73 & 49.92 $\pm$ 0.74 & 73.79 $\pm$ 0.67 & 54.66 $\pm$ 0.77 \\
CE Training    & ConFT (ours)             & ResNet12 & \textbf{76.49 $\pm$ 0.63} & \textbf{64.87 $\pm$ 0.70} & 74.22 $\pm$ 0.71 & 
\textbf{59.23 $\pm$ 0.77}\\ \hline
\multicolumn{1}{l}{} & \multicolumn{1}{l}{}     & \multicolumn{1}{l}{} & \multicolumn{1}{l}{} & \multicolumn{1}{l}{} & \multicolumn{1}{l}{} & \multicolumn{1}{l}{}
\end{tabular}}
\caption{\textbf{Additional Prior Work Comparison.} SCL introduces a novel attention-based spatial contrastive objective for prior learning. While we employ a much simpler cross-entropy objective for prior learning (see CE training \textit{without} ConFT), finetuning the prior embedding with ConFT outperforms SCL significantly in two (CUB and Cars) out of four domains. Our approach yields competitive results for Plantae as well. Further, due to the complementary nature of finetuning, the best performance might be achieved by combining SCL with our ConFT.}
\label{tab:apw}
\end{table*}

In Table \ref{tab:comple}, we validate the complementary effect of our finetuning approach
to a variety of prior learning schemes. Specifically, we compare our simple cross-entropy objective with ProtoNet \cite{Snell2017PrototypicalNF} and ProtoNet with auxiliary self-supervision \cite{Su2020When}. Both of these approaches are based on meta-learning, and were originally proposed for in-domain few-shot classification where base and novel tasks follow the same distribution. Nevertheless, the embeddings thus learnt are readily applicable to cross-domain tasks as well. For the auxiliary self-supervision, we use image rotation as our pretext task. While previous work \cite{Su2020When} has demonstrated the improvement in  in-domain few-shot generalization resulting from rotation based self-supervision, we found that the improvement is marginal in our cross-domain setting (see ProtoNet without finetuning vs. ProtoNet $+$ Rot. without finetuning in Table \ref{tab:comple}), except for when the novel domain is Cars. To obtain these results, we use the official implementation\footnote{\url{https://github.com/cvl-umass/fsl_ssl}} of \cite{Su2020When} with the same hyperparameters (such as loss weighting term) but different backbone. As our pretrained embedding, we trained a ProtoNet model (with auxiliary self-supervision) based on ResNet10 \cite{He2016DeepRL} architecture. Our main observation from Table \ref{tab:comple} is as follows: while better prior learning objectives such as those with auxiliary self-supervision can improve few-shot classification in the novel domains, finetuning with ConFT consistently leads to large improvements over the prior embeddings.

\section{Additional Comparison with Prior Work}
 
In Table \ref{tab:apw}, we report additional comparison with a concurrent work SCL \cite{ouali2020spatial} that introduces  attention-based spatial contrastive objective in the prior-learning phase. For a fair comparison to SCL, we adopt the same backbone based on the ResNet12 architecture which was originally proposed in \cite{tian2020rethink}. While the spatial contrastive objective benefits from larger image resolution ($224 \times 224$), we found it significantly increases the time for finetuning in our case, especially given the larger backbone. So, in this case, we conduct our experiments with a smaller resolution of $84 \times 84$ . Despite the drop in resolution, our finetuning based approach over simple cross-entropy prior learning outperforms the more sophisticated SCL by significant margins in CUB ($7$ points) and Cars ($13$ points). While we attain similar performance in the case of Plantae, we underperform in Places domain. This gap can be understood as a consequence of a stronger SCL based prior embedding for \textit{mini}ImageNet and greater similarity of  the \textit{mini}ImageNet domain to Places as opposed to other novel domains (see Table \ref{tab:PAD}). Nonetheless, our finetuning is complimentary to SCL, and hence we suspect that the best  performance could be achieved by combining it with our ConFT.
 
\section{Meta-Dataset Results}
\begin{table*}
\centering
\begin{tabular}{cccccc} 
\hline
Method                                      & \multicolumn{5}{c}{Target Datasets}                                                                                                                                                                                                                                                                                                                                       \\
                                            & ILSVRC     & Omni          & Aircraft                                    & Birds                                       & DTD    \\ 
\hline
PN \cite{Doersch2020CrossTransformersSF}       &$ 41.87_{\pm0.89} $  &   $ 61.33_{\pm1.13}    $ &   $ 39.40_{\pm0.78}  $   &   $ 65.57_{\pm0.73} $     &   $ 59.06_{\pm0.60} $    \\
CTX \cite{Doersch2020CrossTransformersSF}                &$ 51.70_{\pm0.90} $  & $ 84.24_{\pm0.79}    $   &   $ 62.29_{\pm0.73 }  $   &   $\mathbf{79.38_{\pm0.54}}$ & $ \mathbf{65.86_{\pm0.58}} $ \\
CTX+SC \cite{Doersch2020CrossTransformersSF}     &$ 51.29_{\pm0.89} $  &   $ 86.14_{\pm0.74   } $ &   $ \mathbf{69.74_{\pm0.67 }}                                 $   &   $ 74.85_{\pm0.62       }                           $   &   $ 63.84_{\pm0.62   }                              $    \\
CTX+SC+Aug \cite{Doersch2020CrossTransformersSF} &$ 52.56_{\pm0.86 }$  &   $ 87.53_{\pm0.61  }  $ &   $ 64.28_{\pm0.71  }                                $   &   $ 73.27_{\pm0.63 }                                 $   &   $ 64.72_{\pm0.63  }                                $    \\
ConFT (ours)                                &$ \mathbf{72.07_{\pm 0.71}}$ &   $ \mathbf{98.22 _{\pm 0.17} }$ &   $68.44 _{\pm 0.70} $ &   $ 74.93 _{\pm 0.67}$ &$   63.11 _{\pm 0.70} $\\
\hline
\end{tabular}

\begin{tabular}{cccccc} 
\hline
Method                                      & \multicolumn{5}{c}{Target Dataset}                                                                                                                                                                                                                                                                                                                                       \\
                                            &  QDraw                                      & Fungi      & Flower                                      & Sign                                        & COCO                                         \\ 
\hline
PN \cite{Doersch2020CrossTransformersSF}       & $ 47.86_{\pm0.80}$ &   $ 41.64_{\pm1.02} $    &   $ 83.88_{\pm0.48}$ &   $ 44.84_{\pm0.88}$ &   $ 41.14_{\pm0.82}$\\
CTX \cite{Doersch2020CrossTransformersSF}                &   $ 63.36_{\pm0.73  }                               $&$ 49.43_{\pm0.98 }$   &   $ 92.74_{\pm0.29        }                          $   &   $ 68.31_{\pm0.71    }                              $   &   $ 48.63_{\pm0.79}  $                                 \\
CTX+SC \cite{Doersch2020CrossTransformersSF}     &   $ 64.11_{\pm0.67         }                        $    &   $ 48.87_{\pm0.91} $    &   $ 93.00_{\pm0.30 }                                 $   &   $ 70.62_{\pm0.68        }                          $   &   $ 48.45_{\pm0.83 } $                                 \\
CTX+SC+Aug \cite{Doersch2020CrossTransformersSF} &   $ 66.90_{\pm0.66             }                    $    &   $ 48.22_{\pm0.94} $    &   $ 93.23_{\pm0.28 }                                $   &   $ 78.45_{\pm0.60            }                      $   &   $ 56.61_{\pm0.78} $                                  \\
ConFT (ours)                            &$ \mathbf{80.02 _{\pm 0.6}} $&$\mathbf{50.16_{\pm0.80}}$ &$ \mathbf{94.52 _{\pm 0.29}} $&$ \mathbf{88.22 _{\pm 0.59}} $& $\mathbf{70.73 _{\pm 0.79}}$  \\
\hline
\end{tabular}

\caption{\textbf{Meta-Dataset Results (5-shot). }Cross-domain results of our distractor-aware contrastive finetuning (ConFT) on transfer from ImageNet-only are presented here. The accuracies are averaged over 600 evaluation tasks with $95\%$ confidence intervals. PN: Prototypical Net, SC: SimCLR Episodes. }
\label{tab:metad}
\end{table*} 
In this section, we present the results of our ConFT approach on Meta-Dataset (see Table \ref{tab:metad}). Here, we use an off-the-shelf ResNet18 model\footnote{\url{https://github.com/peymanbateni/simple-cnaps}} pretrained  on  ImageNet-train-split  of  Meta-Dataset  using  just  cross-entropy  objective.  In order to maintain consistency with pretraining, our finetuning operates at a small image resolution of $84\times84$. In this experiments, we keep most of the hyperparameters fixed  across all datasets. In particular, we use a temperature of $0.1$, a distractor batch size of  $128$, and a learning rate of $5e-5$. The early stopping epoch is cross-validated using the meta-validation splits of respective datasets. We obeserve that our approach outperforms the state of the art in 
$7$ out of $10$ datasets and sometimes by a significant margin. This is despite the fact that our input resolution is much smaller compared to $224\times224$ in the state of the art and our approach does \textit{not} benefit from a tansductive setting. Finally, our results reinforce the superiority of simple finetuning over more complex meta-learning frameworks (\eg cross-attention based) even when the domain gap is large. 

\end{document}